# Few-Shot Learning for Mental Disorder Detection: A Continuous Multi-Prompt Engineering Approach with Medical Knowledge Injection

Haoxin Liu (co-first author)
Hefei University of Technology
Email: hxliu@mail.hfut.edu.cn

Wenli Zhang, Ph.D. (co-first author)
Iowa State University
Email: wlzhang@iastate.edu

Jiaheng Xie, Ph.D.
University of Delaware
Email: jxie@udel.edu

Buomsoo (Raymond) Kim, Ph.D.
Iowa State University
Email: rkim0927@iastate.edu

Zhu Zhang, Ph.D.
University of Rhode Island
Email: zhuzhang@uri.edu

Yidong Chai, Ph.D.
Hefei University of Technology
Email: chaiyd@hfut.edu.cn

Sudha Ram, Ph.D.
University of Arizona
McClelland Hall 430J, 1130 E. Helen St.
Tucson, Arizona 85721-0108
Email: ram@eller.arizona.edu

Please send comments to Wenli Zhang at wlzhang@iastate.edu.

# Few-Shot Learning for Mental Disorder Detection: A Continuous Multi-Prompt Engineering Approach with Medical Knowledge Injection

## ABSTRACT

This study harnesses state-of-the-art AI technology for detecting mental disorders through user-generated textual content. Existing studies typically rely on fully supervised machine learning, which presents challenges such as the labor-intensive manual process of annotating extensive training data for each research problem and the need to design specialized deep learning architectures for each task. We propose a novel method to address these challenges by leveraging large language models and continuous multi-prompt engineering, which offers two key advantages: (1) developing personalized prompts that capture each user's unique characteristics and (2) integrating structured medical knowledge into prompts to provide context for disease detection and facilitate predictive modeling. We evaluate our method using three widely prevalent mental disorders as research cases. Our method significantly outperforms existing methods, including feature engineering, architecture engineering, and discrete prompt engineering. Meanwhile, our approach demonstrates success in few-shot learning, i.e., requiring only a minimal number of training examples. Moreover, our method can be generalized to other rare mental disorder detection tasks with few positive labels. In addition to its technical contributions, our method has the potential to enhance the well-being of individuals with mental disorders and offer a cost-effective, accessible alternative for stakeholders beyond traditional mental disorder screening methods.

# 1. INTRODUCTION

Mental disorders are a major global health burden. There are over 150 recognized core mental health conditions (APA 2022) and approximately 1 in 8 people worldwide (970 million), live with a mental disorder (WHO 2022a). Recent studies indicate a significant 25% increase in anxiety and depression, two key mental disorders, since the onset of the COVID-19 pandemic in 2020 (WHO 2022a). More importantly, despite considerable research efforts, mental disorders are difficult to detect for reasons such as the lack of a reliable laboratory test for diagnosis and insufficient behavioral data in electronic health records (EHR) for effective detection (WHO 2022b). Moreover, mental disorders continue to be underdiagnosed due to factors such as lack of awareness of their symptoms, myths and misunderstandings, stigma leading individuals to hide their issues and delay seeking help, and barriers to healthcare access (Patel et al. 2018). Using user-generated content as a supplement to existing mental disorder screening methods is considered a promising approach to combating mental disorders and has far-reaching health and societal implications (Guntuku et al. 2017, Wongkoblap et al. 2017). Information systems (IS) research closely follows this direction and emphasizes the use of user-generated content for mental disorder detection (Chau et al. 2020, D. Zhang et al. 2024, W. Zhang et al. 2024).

Previous work in data-driven healthcare studies demonstrates that patients with chronic diseases, including mental disorders, consistently share their symptoms, life events associated with their conditions, and details of their treatments through user-generated textual content online (Abbasi et al. 2019, Chau et al. 2020, Zhang and Ram 2020). Among the various healthcare studies that leverage user-generated content, research on mental disorders is particularly well-suited for utilizing this type of data because the current diagnosis of mental disorders relies on self-reported symptoms and life events in natural languages (APA 2022); and decades of scientific research have shown that user-generated text can reveal people's psychological states (e.g., emotions, moods) as well as behaviors and activities often associated with mental disorders (Tausczik and Pennebaker 2010, W. Zhang et al. 2024). Hence, using AI to analyze user-generated content holds great potential for enhancing the detection and management of mental disorders by extracting valuable insights from individuals' firsthand experiences (Bardhan et al. 2020). For instance, online platforms can leverage mental disorder detection techniques to develop new services featuring personalized recommendations for users (e.g., encouraging individuals to seek help and treatments, promoting educational content and tools, offering treatment options, and fostering social support). Public administration can employ these techniques to strategically allocate resources to areas with high incidence rates, thereby enhancing the overall effectiveness of chronic disease programs. Policymakers can monitor large-scale user-generated textual content and facilitate the creation of evidence-based policies tailored to the specific needs of different patient cohorts.

However, existing methods on mental disorders through user-generated content show limitations in terms of real-world applicability and generalizability due to the heavy reliance on fully supervised learning. Each mental disorder has unique characteristics and features, and these distinctions could include variations in the symptoms that patients reported, the life events that led to the development or progression of the disease, and the specific treatments that were applied. As a result, researchers have to follow a labor-intensive process to analyze and predict outcomes for these chronic diseases. For instance, it is required to collect and label data, creating a dataset where examples are categorized or identified according to the specific disease they were related to. However, it is difficult to reuse a dataset for a specific chronic disease for another disease using a fully supervised learning model, resulting in the need to create multiple datasets for different mental disorders. Furthermore, a customized machine learning model for each individual disease has to be designed and fine-tuned, which involves meticulously optimizing the algorithms, parameters, and features to make the model effective for that particular disease. This process is highly costly in terms of time, effort, and resources, significantly hampering the applicability of the resulting prediction model.

This study aims to address this research gap by developing a generalizable and adaptable method capable of detecting multiple mental disorders without constructing a large amount of training data or designing a customized model for each disease. With the emergence of LLMs in AI and their remarkable abilities across various downstream tasks, the learning paradigms in NLP-related tasks have evolved from traditional feature engineering and architecture engineering to more advanced learning paradigms, including fine-tuning and prompt engineering (Liu et al. 2023). The foundation for such an approach is grounded in LLMs that have already employed extensive training data, computing power, and algorithmic capabilities to achieve highly promising results across various domains, surpassing the performance of individually-trained, task-specific models. Given these technological advances, leveraging the immense potential of LLMs and prompt engineering for mental disorder detection using user-generated content can minimize the cost of training and developing disease- or problem-specific models. This approach offers a more effective and efficient alternative than traditional methods.

Nevertheless, challenges remain in utilizing prompt engineering and LLMs for mental disorder detection. Most current studies on prompt engineering in IS focus on discrete prompts–how to tweak the natural language used in prompts for downstream tasks. However, we argue that this approach is not the most effective in the context of mental disorder detection: a binary classification task where classification performance is paramount, and human comprehension of the prompt itself is not crucial. Moreover, detecting mental disorders using user-generated content presents two unique challenges. (1) Detecting heterogeneity among various diseases and how each individual presents their conditions poses significant challenges for prompt engineering. Different mental disorders exhibit distinct characteristics, including unique symptoms, risk factors, and treatment approaches, all of which shape the content of user-generated material. Additionally, each patient has a unique persona, characterized by individual patterns, habits, and disease progression, which collectively influence the creation of user-generated content. As a result, user-generated content can be lengthy, noisy, and highly complex, making it challenging for LLMs to efficiently detect various mental disorders at the subject level. This task extends beyond simply identifying explicit mentions of mental disorders and presents significant challenges for LLM models in comprehending nuances, extracting, understanding, and inferring the implicit information related to mental disorders. (2) Incorporating structured medical domain knowledge into prompt engineering to enhance an LLM's predictive performance remains under-explored. Within the medical domain, there is a wealth of medical knowledge that can be closely linked to the content reported in user-generated content and pertains to mental disorders, which can provide significant assistance in employing LLMs for mental disorder detection. However, existing medical knowledge is often organized in tree or network structures (e.g., ontologies, one of the most prevalent forms of domain knowledge). Current discrete prompt methods, which refine questions in natural language, lack effective mechanisms to leverage such structured knowledge.

These two unique challenges in detecting mental disorders using user-generated content motivate us to propose a novel method that aims to: (1) maximize the performance of binary classification for mental disorder detection; (2) account for significant individual differences in both the disorder and the individuals during the prompt engineering process to improve predictive performance; and (3) enhance the effectiveness of LLM predictions by injecting medical knowledge in the form of ontologies during the prompting process. Specifically, our proposed framework utilizes continuous ensemble prompt engineering techniques to interact with LLMs and generate accurate mental disorder detection results. We incorporate prefix-tuning to create personalized prompts tailored to individual patients. Additionally, to account for the unique characteristics of each mental disorder and leverage medical knowledge, we integrate a novel rule-based prompting method that incorporates disease-related medical ontologies.

Our key contributions are twofold. From the healthcare domain perspective, we propose a novel approach using prompt engineering and LLMs for the detection of mental disorders through user-generated textual content and achieve few-shot learning. The key advantage lies in eliminating the need for a substantial amount of labeled training data or customized architecture engineering for each specific disease or research problem. From the methodology perspective, we have two innovations. (1) We propose an ensemble prompt method, synergizing prefix tuning and rule-based prompt engineering to address challenges in healthcare: personalized prompts and medical knowledge injection, which enhance method accuracy and efficacy. (2) We propose a new rule-based prompt method that efficiently tackles complex detection problems, integrating ontology-format domain knowledge, and its design principles can be extended to other problem domains, maximizing the potential of LLMs for real-world problem-solving.

We position our work as computational design science research (Gregor and Hevner 2013, Hevner et al. 2004). In the context of machine learning in IS research (Padmanabhan et al. 2022), our work represents a Type I contribution focused on method development. Specifically, we introduce a new continuous ensemble prompt engineering method for personalized context and medical knowledge injection. Given that mental disorders are one of the major contributors to the overall global disease burden, our approach addresses this societal challenge by providing a tailored machine learning framework along with accompanying algorithms (Padmanabhan et al. 2022). In line with design research pathways for artificial intelligence (Abbasi et al., 2024), our "artifact typologies" are a new "predictive model". The "abstraction spectrum" of our work includes: (1) emphasizing individual differences in the prompt-based prediction process to enhance accuracy, and (2) incorporating existing domain knowledge in ontology format into the prompting process can significantly improve performance. Both contribute valuable "salient design insights" for future research.

Practically, our work has significant implications for mental disorder detection. It provides an accurate detection method that can provide complementary information to existing mental disorder screening procedures. For public health management, our method enables large-scale analyses of a population's mental health beyond what has previously been possible with traditional methods.

## 2. RELATED WORK

Our work aims to leverage user-generated text content for detecting mental disorders, framing it as a binary classification problem. We also propose a novel prompting method that utilizes large LLMs and overcomes the limitations of current mental disorder detection approaches, which often rely heavily on large amounts of labeled training data. We begin by reviewing the evolution of supervised machine learning techniques in natural language processing (NLP) and explaining why prompt-based methods hold promise in this research area. Next, we provide an overview of existing prompting techniques, justifying the motivations behind our research design and emphasizing the novelty of our proposed method.

### 2.1. Paradigms in NLP-related Supervised Machine Learning

Supervised learning, a subcategory of machine learning and AI, has found extensive applications across diverse domains, facilitating tasks such as classifications, detections, and predictions. It is characterized by its use of labeled training datasets to supervise algorithms that produce outcomes accurately. In NLP (i.e., textual content-related machine learning), supervised learning has its paradigms and has evolved through various stages (Figure 1): from feature engineering and architecture engineering to pre-training and fine-tuning, and finally, to pre-training and prompt engineering (Liu et al. 2023).

Until recently, most studies have focused on fully supervised learning. Since fully supervised learning requires a substantial amount of labeled data to train high-performing models, and large-scale labeled data for specific NLP or healthcare-related tasks are limited, researchers have primarily focused on *feature engineering* before the advent of deep learning. Feature engineering involves extracting meaningful features from data using domain knowledge. For instance, Chau et al. (2020) focus on identifying emotional distress in user-generated content by employing a combination of feature extraction, feature selection, rules derived from domain experts, and machine learning classification.

With the emergence of deep learning, which has the capacity to automatically extract features from data without feature engineering, researchers shifted their focus to model *architecture engineering*. These approaches involve designing appropriate deep learning structures to introduce inductive biases into models, facilitating the learning of useful features. A notable work is Yang et al. (2022), in which the authors develop a deep learning architecture for personality detection using user-generated content. Their research design is deliberately crafted to incorporate advanced deep learning architecture engineering, including transfer learning and hierarchical attention network architectures, alongside concepts from relevant psycholinguistic theories.

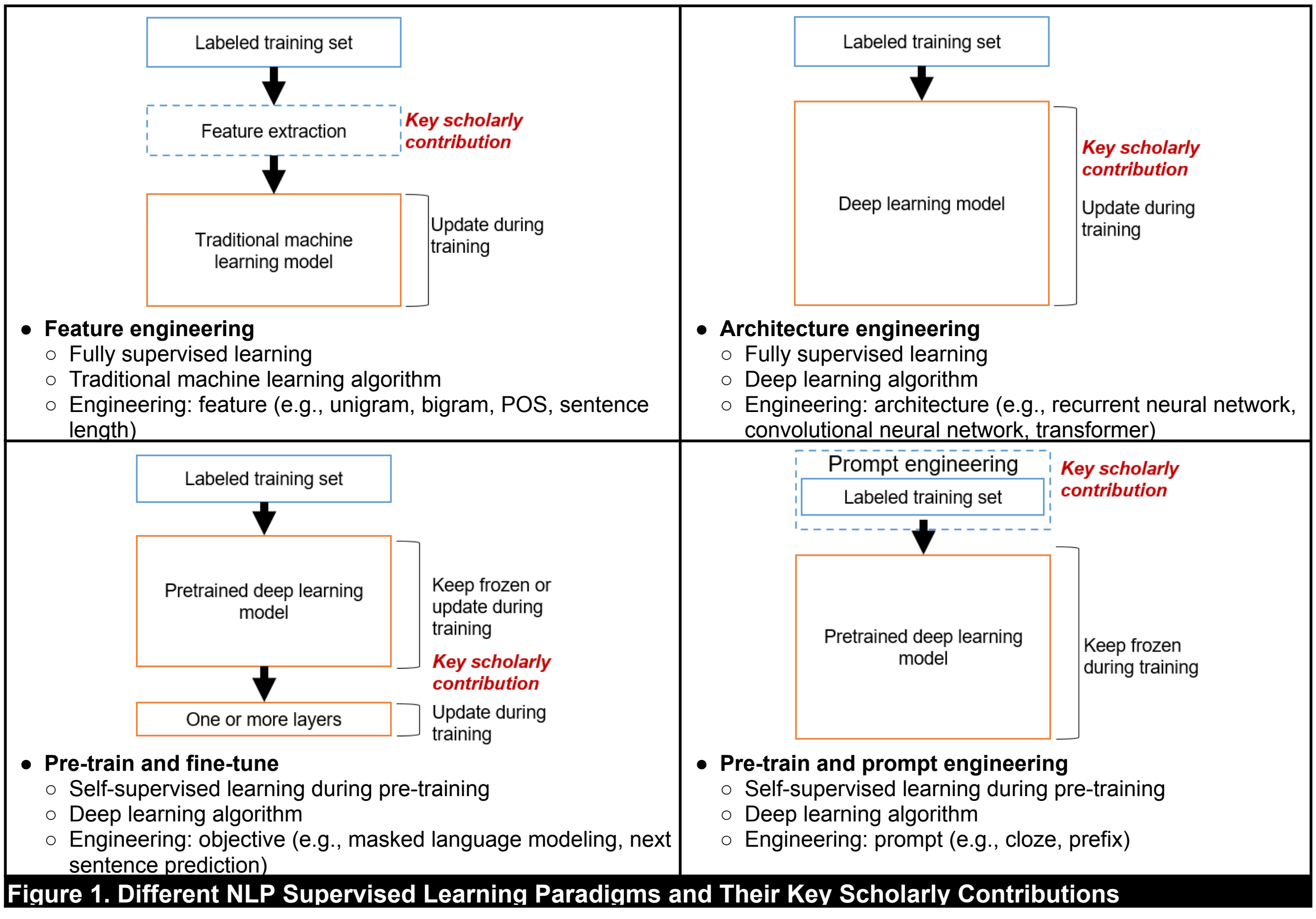

**Figure 1. Different NLP Supervised Learning Paradigms and Their Key Scholarly Contributions**

Since 2018, NLP-related machine learning models transitioned to a new paradigm known as *pre-train and fine-tune* (Devlin et al. 2018), where a fixed architecture language model (e.g.,

BERT, T5, and GPT) can be pre-trained on a massive amount of text data. *Pre-training* typically involves tasks such as completing contextual sentences (e.g., fill-in-the-blank tasks), which do not require expert knowledge and can be directly performed on pre-existing large-scale data (i.e., self-supervised learning). The pre-trained model is then adapted to downstream tasks by *fine-tuning* (i.e., introducing additional parameters). This shift led researchers to focus on *objective engineering*, involving designing better objective functions for both pre-training and fine-tuning tasks (Sanh et al. 2020).

**Table 1. Example of Prompts in NLP Tasks**

| Task | Original input | Prompt | Input to an LLM | Feedback from an LLM |
|---|---|---|---|---|
| Sentiment prediction | I missed the bus today. | • *I felt so* | I missed the bus today. *I felt so [mask].* | The LM fills in the *[mask]* with an emotion word, e.g., *frustrating.* |
| Translation | | • *English:*<br>• *French:* | *English:* I missed the bus today. *French: [mask].* | The LM fills in the *[mask]* with the corresponding French sentence, e.g., *J'ai raté le bus aujourd'hui.* |

During the process of objective engineering, using different prompts to frame the same input (templates $T$ surrounding the input) could facilitate various tasks and improve predictive performance. Table 1 provides examples using discrete prompts (natural language). However, it is important to note that the templates $T$ surrounding the input can also be continuous prompts (i.e., numeric vectors). Subsequently, it was discovered that even using different prompts for the same task can result in a variance in the prediction performance. That is, not only how the language model is trained but also how the prompt is designed can have a significant impact on the performance. Therefore, many researchers have shifted their focus to *prompt engineering*, exploring the design of effective prompts for downstream tasks (Liu et al. 2023).

Although both fine-tuning and prompt engineering leverage the capabilities of LLMs for various downstream prediction tasks, researchers have found that prompt engineering can outperform fine-tuning in terms of predictions. This is because the objectives of pre-training of LLMs (e.g., masked language modeling) and fine-tuning (e.g., binary classification) are not always aligned, making it difficult to fully exploit the knowledge embedded in pre-trained LLMs, which can lead to suboptimal performance on downstream tasks (Wang et al. 2022). Moreover, empirical evidence suggests that prompt learning is computationally efficient, as it is well-suited for zero-shot or few-shot learning scenarios where data is limited (Gao et al. 2021).

### 2.2. Prompt Engineering

We first elucidate the key differences between prompt engineering and other NLP-related supervised learning paradigms. Feature engineering, architecture engineering, and fine-tuning share a common pattern: training a machine learning model to process labeled training examples ($x$, $y$) and predict an output $\widehat{y}$ as $p(\widehat{y}|x)$. In contrast, prompt engineering follows a distinct learning process: using an LLM, it directly models the probability of an outcome $z$ (see Table 1: Feedback from LM). To leverage these models for prediction tasks, the original input $x$ undergoes modification using a template $T$ to create a new input $x'$ (i.e., the prompt). The template $T$ can be either discrete (natural language) or continuous (numeric vectors).

To clarify the operational mechanisms of prompt engineering, we employ a discrete, human-readable prompt as an illustrative example. This new input $x'$ of LLMs has unfilled slots $[mask]$, and the LLM is then utilized to probabilistically fill in the missing information $[mask]$, resulting in $z$, from which, the ultimate output $\widehat{y}$ can be derived through $p(\widehat{y}|z)$. Take the sentiment prediction task in Table 1 as an example. The original input $x$ is the text "*I missed the bus today*" or its vector representation. The corresponding label, denoted as $y$, is "*negative (sentiment)*." In learning paradigms other than prompt engineering, $\widehat{y}$ is directly derived through $P(\widehat{y}|"I\ missed\ the\ bus\ today")$. However, in prompt engineering, researchers first design a prompt denoted as $x'$ (i.e., a new input to an LLM): "*I missed the bus today. I felt so [mask].*" The LLM fills the unfilled slots of $x'$, resulting in $z$, "*frustrating*." The prediction result $\widehat{y}$ is determined by the value of $z$. As $z$ is closely associated with a negative sentiment, the prediction outcome $\widehat{y}$ is consequently classified as "*negative*." The determination of which words are more closely associated with “*negative*” or “*positive*” can be either pre-defined or learned automatically, which is referred to as the verbalizer $V$.

**Table 2. Representative Prompt Engineering Methods and Comparison with Our Method**

(a) Classification based on Prompt Design

| Cat. | Shape of prompts | | Manual/automated prompts | | Discrete/continuous prompts | | Static/dynamic prompts | |
|---|---|---|---|---|---|---|---|---|
| | Cloze | Prefix | Manual | Automated | Discrete | Continuous | Static | Dynamic |
| LAMA (Petroni et al. 2019) | ✔ | | ✔ | | | | ✔ | |
| TemplateNER (Cui et al. 2021) | ✔ | | ✔ | | | | ✔ | |
| GPT-3 (Brown et al. 2020) | | | ✔ | | | | ✔ | |
| Prefix-Tuning (Li and Liang 2021) | | ✔ | | | | ✔ | | ✔ |
| Prompt tuning (Lester et al. 2021) | | ✔ | | | | ✔ | | ✔ |
| AutoPrompt (Shin et al. 2020) | | | | ✔ | ✔ | | | |
| **Ours** | ✔ | ✔ | | ✔ | ✔ | ✔ | | ✔ |

(b) Classification based on Multi-prompt Learning

| Cat. | Prompt Ensemble | Prompt Augmentation | Prompt Composition | Prompt Decomposition |
|---|---|---|---|---|
| BARTScore (Yuan et al. 2021) | ✔ | | | |
| GPT-3 (Brown et al. 2020) | | ✔ | | |
| PTR (Han et al. 2022) | | | ✔ | |
| TemplateNER (Cui et al. 2021) | | | | ✔ |
| **Ours** | ✔ | | ✔ | |

*Note:*

- Cloze prompt: A type of prompt designed to generate responses that fill in the blanks or complete a sentence.
- Prefix prompt: A type of prompt where a specific instruction or context is provided at the beginning of the input to guide the language model’s response.
- Manual prompt: Manually create intuitive templates based on human introspection.
- Automated prompt: Automate the template design process.
- Static prompt: Using the same prompt for each input.
- Dynamic prompt: Generating a custom template for each input.
- Discrete prompt (hard prompts): A prompt described in a discrete space, usually corresponding to natural language phrases.
- Continuous prompts (soft prompts): Perform prompting directly in the embedding space of the model: (1) relax the constraint that the embeddings of template words be the embeddings of natural language (e.g., English) words, and (2) remove the restriction that the template is parameterized by the pre-trained LLM’s parameters.

The objective of prompt engineering is to develop a prompting function, denoted as $x' = f_{prompt}(x)$, to achieve optimal performance in the subsequent task. Prompt engineering can significantly enhance the efficiency and effectiveness of the prediction process since it enables LLMs to undergo pre-training on vast amounts of pre-existing textual data. Moreover, by defining $f_{prompt}(x)$, the model can facilitate few-shot or even zero-shot learning, seamlessly adapting to new scenarios with minimal or no labeled data.

As the literature underscores, the design of a prompt can have substantial influence on the overall performance of a prompt-based method (Liu et al. 2023). Therefore, various prompt engineering methods have been proposed, which can be categorized based on the shape of prompts, manual/automated prompts, discrete/continuous prompts, and static/dynamic prompts, each with distinct characteristics and associated pros and cons. The choice of prompt engineering method depends on both the task at hand and the specific LLMs employed to address the task. We provide a summary of representative studies in Table 2. Recently, many studies have highlighted the significant improvement in the effectiveness of prompt engineering methods through the utilization of multiple prompts—a concept known as multi-prompt engineering. Several key strategies for multi-prompt learning have been identified, including prompt ensembling, prompt augmentation, prompt composition, and prompt decomposition (Liu et al. 2023).

Although prompt engineering has shown significant potential among different tasks and scenarios, many challenges remain (Liu et al. 2023). Two of the most significant technical challenges in this field are as follows. (1) Prompt design for complex tasks: the formulation and design of prompts for complex tasks are not straightforward (Liu et al. 2023). Particularly, prompt design in mental disorder detection using textual data is under-explored. Each patient possesses unique characteristics and patterns, including but not limited to, linguistic styles (such as a tendency to complain, convey setbacks; or a tendency to endure, face challenges positively), habits of using social media, and the extent to which one is willing to openly discuss their own illnesses, a unique course of progression in their illness, and so on. Moreover, different types of mental disorders exhibit distinct (but sometimes similar) symptoms, risk factors, and treatments. (2) Prompt engineering with structured domain knowledge (Han et al. 2022): in many NLP tasks, inputs may exhibit various structures (e.g., syntax trees or relational structures from relationship extraction); effectively expressing these structures in prompt engineering poses a significant challenge. In the realm of mental disorder management, chronic disease management, and healthcare in general, a substantial volume of medical knowledge exists in structured formats (e.g., ontologies, which are tree or network structures). Leveraging the existing domain knowledge can greatly enhance disease detection using textual data. However, this domain remains largely underexplored, presenting a potentially crucial and interesting avenue for research. These two challenges represent the core issues that this work aims to address. In the following sections, we will explore our proposed solutions within our research context.

### 2.2.1. Mental Disorder Detection and Continuous Prompt Engineering

One method of classifying prompts is to categorize them as either continuous or discrete. A discrete prompt modifies the input to an LLM using natural language. In contrast, a continuous prompt operates directly in the model's embedding space, allowing it to (1) relax the constraint that template embeddings $T$ must correspond to natural language (e.g., English) words and (2) eliminate the restriction that the template $T$ is parameterized by the pre-trained LLM's parameters.

As mentioned earlier, this study focuses on detecting mental disorders using user-generated content. The problem is a binary classification task. We argue that continuous prompts are more effective than discrete ones for our research problem for the following reasons: (1) Enhanced expressiveness: continuous prompts leverage high-dimensional vector representations, enabling the model to learn latent semantic features tailored to the classification task (Lester et al. 2021). Unlike discrete prompts, continuous prompts capture nonlinear relationships that rigid textual templates (discrete prompt) often miss (Liu et al. 2024). (2) Flexibility via optimization: continuous prompts consist of trainable parameters optimized through backpropagation, allowing them to align directly with the prediction task's loss function (Lester et al. 2021). In contrast, discrete prompts rely on heuristic tuning, which may lead to misalignment with the LLMs' internal representations and the objectives of the prediction task (Shin et al. 2020). (3) Mitigating ambiguity in classifications: binary tasks, such as mental disorder detection (positive/negative), require probabilistic boundaries. Continuous prompts excel in these domains by providing probabilities as feedback from LLMs. In our work, we leverage these probabilities for result interpretation, which can offer significant benefits to stakeholders (see Section 4.5). In contrast, discrete prompts impose rigid mappings, only yielding binary outcomes (benign/malignant) (Shin et al. 2020). (4) Computational efficiency and performance improvement: although discrete prompts may seem straightforward because they are easily understandable by humans, their heuristic tuning (e.g., words tweaking for better performance) can be time-consuming. In contrast, continuous prompts automate the tuning process with minimal computational overhead (add minimal trainable parameters, e.g., prefix embeddings, avoiding full model fine-tuning). This is critical for binary tasks with limited labeled data (Lester et al. 2021).

Continuous prompts have one primary limitation: they are typically represented as high-dimensional numerical vectors, which makes it difficult to directly understand or interpret their specific meanings. This contrasts with the more intuitive nature of discrete prompts. However, these limitations are not critical in the context of our research–binary classification. The understandability of the prompt per se (the input) is not the ultimate goal. Our goal is to predict mental disorders (the output). Even with continuous prompts, we can still offer explainable insights for this prediction outcome, which we will show in Section 4.5.

Moreover, there are two key research challenges in our research problem: (1) addressing heterogeneity among individuals and disorders, and (2) incorporating structured medical knowledge into prompts to enhance LLM prediction. We aim to tackle this research problem by enhancing the performance of LLMs in binary mental disorder detection tasks through a prompt ensemble approach that combines two continuous prompt engineering techniques: prefix tuning and rule-based prompting.

#### 2.2.2. Prefix Tuning for Personalized Prompts

Prefix tuning is a continuous prompt method that optimizes a sequence of trainable vectors prepended to the input embeddings (Li and Liang 2021). The underlying intuition of this method lies in the idea that by providing an appropriate *context* to an LLM, which can influence the encoding of $x$ and direct the LLM on what information to extract from $x$. Therefore, the *context* can guide the LLM to effectively solve downstream tasks. Nevertheless, it is not clear whether such a *context* exists or how to identify such a *context* for each individual $x$. Therefore, the authors propose the prefix tuning method to automatically optimize continuous prefixes for inputs as the *context*. Formally, for a training example $(x, y)$, they define

$$f_{prompt} = [Prefix;\ x;\ Prefix';\ y] \tag{1}$$

where $Prefix$ and $Prefix'$ are placeholders for values associated with the training example $(x, y)$. The $Prefix$ and $Prefix'$ for all training examples consist of a trainable matrix $P_{\theta}[i,:]$, where $i \in P_{idx}$ and $P_{idx}$ denotes the sequence of prefix indices. Therefore, the feedback from LLM is

$$z_i = \begin{cases} P_{\theta}[i,:], & \text{if } i \in P_{idx}, \\ LM_{\phi}\left(f_{prompt}(x)_i, z_{<i}\right), & \text{o.w.} \end{cases} \tag{2}$$

$z_i$ represents a function of the trainable $P_{\theta}$ (of dimension $\left|P_{idx}\right| \times dim(z_i)$). When $i \in P_{idx}$, $z_i$ directly copies from $P_{\theta}$; when $i \notin P_{idx}$, $z_i$ still depends on $P_{\theta}$, as it is the prefix context and subsequent feedback from $LM_{\phi}$ relies on the activations of the preceding feedback. Empirically, they reparametrize the $P_{\theta}[i,:]$ using matrix $P'_{\theta}[i,:]$ (of dimension $\left|P_{idx}\right| \times k$, where $k$ is a hyperparameter) using a feedforward neural network for stable training: $P_{\theta}[i,:] = MLP_{\theta}(P'_{\theta}[i,:])$. The learning goal is

$$max_{\phi} log\, p_{\phi(y|x)} = \Sigma_{i \in y_{idx}} log\, p_{\phi}\left(f_{prompt}(x, y)_i | z_{<i}\right) \tag{3}$$

where $p_{\phi}$ is an LLM distribution, $\phi$ is the LLM parameters that are fixed.

One significant advantage of this method is that it provides flexibility and adaptability to individual input $x$. Given that the prepended vectors (i.e., $Prefix$ and $Prefix'$) are automatically updated during training, each prefix vector (i.e., $P_{\theta}[i,:]$) is customized for individual input $x$ simultaneously. We exploit this feature of prefix tuning to generate personalized prompts for user textual data. In the context of mental disorder detection using user-generated content, it is desirable to provide a distinct prompt for each user for optimal performance since different users have unique characteristics and underlying patterns. Thus, prefix tuning represents a promising research direction for developing continuous prompts optimized for each individual, thereby enhancing the performance of mental disorder detection. Specifically, our intentions are twofold: (1) to refine the design of $f_{prompt}$ (Eq. 1) to better fulfill the role of a personalized prompt tailored to individual users for mental disorder detection; and (2) to seamlessly integrate the learning objective of prefix tuning (Eq. 3) with other prompt learning goals through a multi-prompt approach (i.e., prompt ensemble) to more effectively address the challenges associated with mental disorder detection.

### 2.2.3. Knowledge Injection Through Rule-based Prompts

The rule-based prompt is another continuous prompt method that incorporates logical rules (e.g., "If the text contains word *X*, it likely belongs to class *Y*") into the prompt tuning process. It uses continuous prompts as its foundation while constraining their optimization through rule-based losses (Han et al. 2022). Rule-based prompt is proposed to address the limitations of other widely-used prompt engineering methods in addressing complex text classification tasks: (1) manual prompt design is both laborious and prone to errors, (2) and for auto-generated prompts, the validation of the efficacy is a resource-intensive and time-consuming process.

The essence of the rule-based prompt to solve challenging classification tasks is threefold. First, for a highly challenging text classification problem (i.e., given $(x, y)$ and predict $p(\widehat{y}|x)$), rule-based prompt breaks down the classification question into several simpler sub-classification tasks, namely, breaking down $p(y|x)$ to $p(y^1|x)... p(y^f|x)... p(y^k|x)$, where $k$ indicates the number of subtasks. Then, the rule-based prompt method incorporates logical rules to compose task-specific prompts with several simpler *sub-prompts* and accomplish the complex classification task. Formally, for each sub-classification task $p(y^f|x)$, the rule-based prompt method sets a template $T^f(x)$ and a set of verbalizer words $V^f = \{v_1,..., v_n\}$. The template $T^f(x)$ and verbalizer $V^f$ constitute the prompting function $f^f_{prompt}(x)$. The logical rule is defined as

$$p(y^1|x) \wedge p(y^2|x)... p(y^{f-1}|x) \wedge p(y^f|x)... p(y^{k-1}|x) \wedge p(y^k|x) \rightarrow p(y|x) \quad (4)$$

Second, the rule-based prompt method incorporates prior knowledge for each sub-classification task, reducing the laborious and error-prone nature of manual prompt construction and mitigating the uncertainties associated with auto-generated prompts. Formally, when constructing each sub-prompt $f^{f}_{prompt}(x)$ for each sub-classification task $p(y^{f}|x)$, prior knowledge can be injected in both the design of $T^{f}(x)$ and verbalizer $V^{f}$ to facilitate the prediction and performance of $p(y^{f}|x)$. For instance, consider a classical sub-classification problem in named entity recognition. Let $T^{f}(x) = "x\ is\ the\ [mask]\ entity"$ and $V^{f} = \{"person", "organization",...\}$. Since this sub-task involves named entity recognition, the templates and verbalizers can be meticulously customized to assist an LLM in accurately identifying the entity category. For a classical relation prediction problem, let $T^{f}(x) = "x\ entity_{1}\ [mask]\ entity_{2}"$ and $V^{f} = \{"was\ born\ in", "is\ parent\ of",...\}$. Again, the templates and verbalizers for this sub-classification problem can be tailored to assist an LLM in completing a relation prediction task.

Lastly, the rule-based prompt method composes sub-prompts of various sub-problems into a complete task-specific prompt,

$$f_{prompt}(x) = \begin{cases} T(x) = [T^{1}(x);...; T^{f}(x);...; T^{k}(x)], \\ V[mask]_{1} = \{v^{1}_{1}, v^{1}_{2},...\},..., V[mask]_{2} = \{v^{f}_{1}, v^{f}_{2},...\},..., V[mask]_{k} = \{v^{k}_{1}, v^{k}_{2},...\}. \end{cases} \quad (5)$$

where $[\cdot; \cdot; \cdot]$ is the aggregation function of sub-templates. The learning objective of the rule-based prompt method is

$$max_{\phi} p_{\phi(y|x)} = log \prod_{f=1}^{r} p_{\phi}\left([mask]_{f} = LM_{\phi}(y)|T(x)\right) \quad (6)$$

where $r$ is the number of masked positions in $T(x)$ and $[mask]_{f} = LM_{\phi}(y)$ is to map the class $y$ to the set of label words $V[mask]_{f}$.

In our research context, predicting whether an individual has a specific mental disorder by directly utilizing an LLM and ultra-long user-generated content as inputs (since the task is at the individual level) presents a significant challenge. As mentioned, various mental disorders exhibit distinct or sometimes similar symptoms, risk factors, and treatments. Therefore, the rule-based prompt method is an efficient way to design sub-prompts to capture different aspects of mental disorders (e.g., symptoms, risk factors, and treatments) to simplify the detection task using user-generated content. Furthermore, the rule-based prompt method is an ideal method to incorporate the existing domain knowledge which is widely available and essential in mental disorder diagnosis and healthcare. Hence, in this study, we attempt to encode and incorporate existing medical knowledge by proposing a new rule-based prompt engineering method for improved mental disorder detection performance. Specifically, our key innovations include: (1) modifying the logic rules implemented in the original method (Eq. 4) and the learning goal of the rule-based prompt method (Eq. 6) to transfer it to the mental disorder detection task; (2) exploring an effective mechanism to inject existing medical knowledge of mental disorder detection in the rule-based prompt engineering process (Eq. 5), and (3) seamlessly integrating the learning objective of the rule-based prompt method (Eq. 6) with other prompt learning goals through a multi-prompt approach (i.e., prompt ensemble and prompt composition) to more effectively address the challenges associated with mental disorder detection.

### 2.3. Key Novelties

From the perspective of design science, we make three technical contributions with our main IT artifact developed for mental disorder detection using textual data. First, we present a novel framework grounded in LLMs and prompt engineering, facilitating the few-shot detection of multiple mental disorders through user-generated text content. Notably, this framework confers a significant advantage by obviating the necessity for an extensive volume of labeled training data or the intricate engineering of customized architectures for each distinct disease or research problem. The proposed framework can be extended to tasks related to detecting other mental disorders and chronic diseases, especially those exhibiting discernible characteristics within user-generated textual content. Second, within our framework, we propose a multi-prompt engineering approach, effectively synergizing various continuous prompt engineering techniques, including prefix tuning and rule-based prompt engineering. This strategic amalgamation is specifically tailored to address the unique technical challenges within the healthcare domain. It involves the utilization of personalized prompts and the integration of existing medical domain knowledge, thereby markedly enhancing the accuracy and efficacy of our method. Third, as an integral component of our framework, we propose a new rule-based prompt engineering method, adept at efficiently dissecting complex textual content-based detection problems. This method seamlessly integrates domain knowledge existing in the ontology format—one of the widely adopted formats for domain knowledge. The design principle extends to other research problems necessitating the decomposition of challenging tasks and maximizes the utilization of LLM's potential to address real-world challenges.

## 3. RESEARCH DESIGN

**Figure 2. Research Design**

In this study, we introduce a novel multi-prompt engineering method for detecting mental disorders through user-generated textual content. The innovative design of our multi-prompt engineering method aims to tackle two technical challenges: (1) personalized prompts for individual users and each mental disorder, capturing the unique characteristics and underlying patterns of each user, and (2) integrating prompts with structured medical knowledge to contextualize the task, which instructs the LLMs on the learning objectives and operationalizes prediction goals. Subsequently, the outcomes of the prompts serve as the input for an LLM, which determines whether the targeted user exhibits signs of a mental disorder. The flowchart of our method is shown in Figure 2.

### 3.1. Problem Formulation

We focus on user-generated textual content on online platforms (e.g., Reddit, Twitter, etc.), which potentially encompasses each user's self-reported information relevant to mental disorder detection. Each textual post is denoted as $x_i$. We collect data from a user base $U$ with $l$ users. For a given period of time, we observe the user-generated content of the focal user $u \in U$ from $N$ text posts, denoted by $x_u = \left(x_1, x_2, \ldots, x_{N_u}\right)$, ordered in time, and $N_u = \{1, 2, 3, \ldots\}$ as each user has an arbitrary number of posts. Each user $u$ may suffer from one or more mental disorders in a disease set $D = \left\{d_1, \ldots, d_j, \ldots, d_N\right\}$. For each disease $d_j$, an ontology $O_j$ can be constructed to depict the symptoms, risk factors, and treatments of the disease $d_j$. Given a user's text posts $\left(x_1, x_2, \ldots, x_{N_u}\right)$ and a target disease $d_j$, we aim to design a new multi-prompt function $f_{prompt}\left(x_1, x_2, \ldots, x_{N_u}\right)$ to address two technique challenges: personalized prompts and prompts integrated with medical knowledge $O_j$. As the foundation of our approach, we build upon LLMs, denoted as $LM_\Phi$, with parameters $\Phi$. The prediction outcome is $y_{d_j} = \{0, 1\}$, where 1 suggests that the focal user $u$ suffers or will suffer from the target disease $d_j$. Formally, the mental disorder $d_j$ detection problem is a binary probabilistic classification problem (Eq. 7), which applies to all diseases in $D$.

$$\widehat{y_{d_j}} = argmax\, p\left(y_{d_j} | LM_\Phi\left(f_{prompt}\left(x_1, x_2, \ldots, x_{N_u}\right)\right)\right) \quad (7)$$

### 3.2. Multi-prompt Engineering with Personalization and Knowledge Injection

As outlined in the literature review, the objective of the prompt engineering function, $f_{prompt}\left(x_1, x_2, ..., x_{N_u}\right)$, is to leverage the capabilities of LLMs while streamlining the complexity of disease- or problem-specific model design. The key scholarly contribution of this study lies in the development of $f_{prompt}\left(x_1, x_2, ..., x_{N_u}\right)$.

#### 3.2.1. Automated Continuous Dynamic Prefix Tuning for Personalized Prompts

When patients describe their experience on social media, their patterns and persona[1] are distinct from each other. This motivates us to design a personalized prompt for each user for the mental disorder detection problem. We leverage the attributes of prefix tuning (Li and Liang 2021), where each prefix vector is customized for individual input simultaneously, and adapt it to our multi-prompt method to achieve this goal. Specifically, we designate a one-dimensional vector $v$ of length $k$ for each user $u$,

$$f_{prompt_prefix}(x) = [v \oplus x] \tag{8}$$

where $\oplus$ denotes concatenation and $k$ is a hyper-parameter. In the user base $U$ with $l$ users, a trainable matrix $P$ (of dimension $l \times (k + LM_{\phi}_tokenizing(x))$) will be parametrized using a feedforward neural network, $P_{\theta} = MLP_{\theta}$, during training, where θ is the trainable parameters of the $MLP$ and is used to create unseen users' $f_{prompt_prefix}(x)$. Each row of $P$ can be trained (i.e., reparameterized) simultaneously during the training process to reflect each user's unique characteristics, allowing for personalized prompts. Consequently, the expected feedback from LLM is

$$z_i = \begin{cases} P_{\theta}[:k], & \text{if } i \le k, \\ LM_{\phi}\left(f_{prompt}(x)_i, z_{<i}\right), & \text{o.w.} \end{cases} \tag{9}$$

[1] Including but not limited to, linguistic styles (such as a tendency to complain, convey setbacks; or a tendency to endure, face challenges positively), habits of using social media, and the extent to which one is willing to openly discuss their own illnesses, a unique course of progression in their illness, and so on.

where $i$ is the $i$-th digits of $z$. When $i \leq k$, $z_i$ directly copies from $P_\theta$; when $i > k$, $z_i$ still depends on $P_\theta$, as it is the prefix context and subsequent feedback of our muti-prompt function $f_{prompt}(x)_i$ (section 3.2.3) relies on the activations of the preceding feedback.

### 3.2.2. A New Rule-Based Prompt for Injecting Structural Medical Knowledge

The mental disorder detection task is at the subject level using all the posts in a time period. Therefore, compared to other NLP tasks, the unique challenge of this task lies in the input to a machine learning prediction model, $x = \left(x_1, x_2, ..., x_{N_u}\right)$, which contains ultra-long user-generated unstructured text content with variable lengths (see section 4.1 Table 4, around 1,583,227 ~ 44,077,018 tokens per user). However, extracting valid information from these data is challenging for traditional machine learning models that rely on feature engineering. On the other hand, conventional end-to-end deep learning models may not be able to remember and learn from ultra-long dependencies from $x = \left(x_1, x_2, ..., x_{N_u}\right)$ (e.g., RNNs) or be constrained by a fixed sequence length established during training (e.g., Transformers), making handling a large amount of arbitrary number text posts challenging.

For standard prompt engineering methods, even with the assistance of highly potent LLMs, it remains a challenging task. This is due to (1) the presence of a significant amount of noise in the user-generated content (e.g., the text content can be unrelated to mental disorders, or similar symptoms shared across various mental disorders), making the prediction task difficult. (2) Considering the limited memory capacity of LLMs based on the number of parameters, most LLMs are insufficient to handle the extensive input required for user-level mental disorder detection. Even if the LLMs can handle such inputs, these models typically have a large number of parameters, imposing significant costs in applications. (3) If we employ a method, such as a discrete prompt that utilizes natural language and expects only binary outcomes (e.g., 1/0), it becomes challenging for stakeholders to determine which specific part of the user-generated content leads the LLM to conclude that the subject has a particular mental disorder. Motivated by these challenges, we propose a new rule-based prompt engineering method with three design principles, differing from existing approaches.

**Design principle 1.** Instead of using $x = \left(x_1, x_2, ..., x_{N_u}\right)$ as the input, we expand the individual elements $x_i$ within $x$ and concatenate them into a long list of tokens, $x = \{t_1, t_2, ..., t_{|x|}\}$, where $t_j$ indicates a single token and $|x|$ is the magnitude of $x$. We then design a sliding window $\bar{x}$ and assume $m$ sliding windows in total. Therefore, a sliding window $\bar{x}$ of size $w$ can be represented as $x[i:\ i + w]$, for $i = \{0, w, 2w, ..., (m - 1)w\}$. The number $m$ and the size $w$ of moving windows are hyperparameters correlated and constrained by $|x|$. Empirically, let $|\bar{x}| \ll |x|$ and the content in the sliding window $\bar{x} = \{t_1, t_2, ..., t_{|\bar{x}|}\}$ is used as the new input of $f_{prompt}(\cdot)$ and $LM_{\phi}$, significantly reducing the negative impacts caused by ultra-long sequences in user-level disease detection tasks.

Another significant advantage of this approach is that the $LM_{\phi}$ returns a probability for each $\bar{x}$, which reflects Eq. 7, i.e., the probability that the LLM considers $\bar{x}$ to be related to a mental disorder. In Section 4.5, we leverage this feature to interpret our prediction results. Additionally, through Design principles 2 and 3 discussed below, we can further determine which aspects of existing medical knowledge are related to $\bar{x}$, thus facilitating stakeholders in utilizing our method and results.

**Design principle 2.** Instead of directly relying on $LM_{\phi}$ to determine whether a user $u$ has a mental disorder, we can break down this task into three sub-tasks: assessing whether a focal user $u$ exhibits: (1) symptoms, such as anxiety, fatigue, low mood, reduced self-esteem, change in appetite or sleep, suicide attempt, etc. (APA 2022, Martin et al. 2006, Rush et al. 2003); (2) major life event changes, such as divorce, body shape change, violence, abuse, drug or alcohol use, and so on (Beck and Alford 2014); and (3) treatments, such as medication, therapy, or a combination of these two (Beck and Alford 2014). The construction of this ontology is supported by medical literature and has been rigorously validated through quantitative methods and expert evaluation, as detailed in Appendix 1.

We decompose the task of detecting subject-level mental disorders into three subtasks, relying on the following assumptions: if the subject $u$ displays an increased number of symptoms, self-reports a greater number of life events that may cause or exacerbate disease $d_j$, or discusses the use of treatments associated with the disease $d_j$, the accumulation of such mentions suggests a higher likelihood that the subject $u$ currently suffers from or will suffer from the target disease $d_j$.

Meanwhile, the reason for decomposing the task into these three subtasks is that, in user-generated text, these three aspects are often self-reported by users with mental disorders and are detectable (Coppersmith et al. 2015, Nadeem 2016, W. Zhang et al. 2024). It is worth noting that there are other indicators for detecting mental disorders, such as family history, genetics, and poor nutrition. However, since our research context revolves around user-generated text content, we concentrate on factors that are detectable.

We aim to design an inclusive method that can (1) identify users with easily noticeable signs of disease $d_j$ who are currently undergoing treatment and (2) provide early detection for users at risk of disease $d_j$ in the future, which includes detecting symptoms and life events that may exacerbate depression. In the *disease detection* task, treatment entities play a significant role. This is because, within the population of individuals discussing disease $d_j$, there is a subset of users who have already received clinical diagnoses and have undergone various treatments. When a user openly discusses treatments of disease $d_j$, it strongly indicates that the user likely has disease $d_j$. To ensure the comprehensiveness of our mental disorder detection method, which aims to identify as many patients as possible, treatment-related entities can be highly effective. On the other hand, when it comes to the *early disease detection task*, our method leverages other disease-related factors for early detection, including symptoms and major life event changes.

Formally, we define the following logic rule:

$$p(y_{d_j}^{symptom} | \bar{x}) \vee p(y_{d_j}^{life_event} | \bar{x}) \vee p(y_{d_j}^{treatment} | \bar{x}) \rightarrow p(y_{d_j} | \bar{x}) \quad (10)$$

where $\vee$ is the logical connective "or". The logical "or" ($\vee$) is inclusive, meaning that at least one of the $p(y_{d_j}^{f} | \bar{x})$, where $f \in \{symptom,\ life_event,\ treatment\}$ must be true for the compound proposition, $p(y| \bar{x}) = 1$, to be true. Practically, $p(y_{d_j}^{f} | \bar{x})$ represents the probability feedback from $LM_{\phi}$, indicating whether $LM_{\phi}$ judges $\bar{x}$ to be associated with the sub-task $f$. In actual calculations, for a user $u$, if there are more $\bar{x}$ within $x$ are determined by $LM_{\phi}$ that they are "symptom" *or* "life event" *or* "treatment" of mental disorder $d_j$, there is a higher probability that our framework will predict a correspondingly higher probability that this focal user $u$ has mental disorder $d_j$.

**Design principle 3.** Having $LM_{\phi}$ directly determine whether $\bar{x}$ is a "symptom," "life event," or "treatment" of a mental disorder $d_j$ remains a challenging task as the features of $d_j$ can be highly specific and complex, but sometimes, certain features of mental disorders can be remarkably similar. For instance, feelings of excessive guilt or self-blame are often linked with depression but can also manifest in other disorders, including PTSD, anorexia, and self-harm. Accurately distinguishing between different mental disorders and providing effective interventions is crucial. For depression, it is important to create a friendly environment and offer information related to treatment. For PTSD, it is essential to avoid triggering information related to trauma. In the case of anorexia nervosa and self-harm, users may be at risk of life-threatening situations, necessitating more immediate help and intervention.

We can further enhance the performance of $LM_{\phi}$ by employing prompt engineering to clearly instruct $LM_{\phi}$ on the specific characteristics of the $\{symptom,\ life_event,\ treatment\}$ associated with $d_j$. It is noteworthy that the specificity of these three aspects in various mental disorders is well-documented in the medical literature. If appropriately integrated, such existing medical knowledge can significantly alleviate the challenges faced by $LM_{\phi}$ and predictive models in detecting mental disorders at the user level using user-generated text content. The injection of medical knowledge into prompt design, therefore, is a significant and promising direction.

Accordingly, to leverage medical domain knowledge for mental disorder detection, we adhere to previous studies and adopt the established mental disorder ontology $O_j$ for each disease $d_j$ that explicitly explains the terminologies used in disease $d_j$'s diagnosis and treatments (W. Zhang et al. 2024). The ontology $O_j$ focuses on specific aspects of disease $d_j$, particularly the medical terminologies used in diagnosing disease $d_j$ that are possible to detect from user-generated textual content, formally denoted as $O_{j_f} \in \{symptom,\ life\ event,\ treatment\}$. The purpose of the ontology is to facilitate the detection of symptoms, major life events, and treatments from user-generated text content. Based on the extensive literature review (APA 2022, Beck and Alford 2014, Martin et al. 2006, Rush et al. 2003), a list of concepts, $o_{j_j}$, related to $d_j$ diagnosis (e.g., dejected mood, self-blame, fatal illness, psychotherapy, etc.) is compiled. Next, we organize the terminologies $o_{j_j}$ into three classes for each disease $d_j$: symptom ($O_{j_symptom}$, a collection of symptoms), life event ($O_{j_life\ event}$, a collection of major life event changes that may cause or exacerbate $d_j$), or treatment ($O_{j_treatment}$, medications and therapies). Meanwhile, we determine the relationships between terminologies $o_{j_j}$ and classes $O_{j_k}$ as $o_{j_j} : relation\ O_{j_k}$ (e.g., for "depression" and one of its symptoms "dejected mood", $o_{depression_{dejected\ mood}} : is\ a\ O_{depression_symptoms}$).

Adhering to the three design principles, we formulate our new rule-based prompting function as follows:

$$f_{prompt_rule}(\bar{x}) = \begin{cases} T_{d_j}(\bar{x}) = [T_{d_j}^{symptom}(\bar{x}); T_{d_j}^{life_event}(\bar{x}); T_{d_j}^{treatment}(\bar{x})], \\ V_{d_j}[mask]_{symptom} = \{o_{j_j}\}, o_{j_j} \in O_{j_symptom}, \text{ when } p(y_{d_j}^{symptom} = 1|\bar{x}), \\ V_{d_j}[mask]_{life_event} = \{o_{j_j}\}, o_{j_j} \in O_{j_life_event}, \text{ when } p(y_{d_j}^{life_event} = 1|\bar{x}), \\ V_{d_j}[mask]_{treatment} = \{o_{j_j}\}, o_{j_j} \in O_{j_treatment}, \text{ when } p(y_{d_j}^{treatment} = 1|\bar{x}). \end{cases} \quad (11)$$

where $T^f(\bar{x}) = "\bar{x}\ :relation\ [mask]_f\ of\ \{d_j\}\ f"$, $V_{d_j}[mask]_f = \{o_{j_j}\}$ denotes all the concept in the ontology $O_j$ that belongs to class $f$, and $f \in \{symptom,\ life_event,\ treatment\}$.

Existing medical knowledge, represented as ontology $O_j$, is injected into $f_{prompt_rule}(\bar{x})$ in two ways, and aids $f_{prompt_rule}(\bar{x})$ to instruct the $LM_\phi$ to better accomplish the user-level mental disorder detection task: (1) the relation ($:relation$) between concept $o_{j_j}$ and concept class $O_{j_k}$ is injected into the prompt template $T_{d_j}(\bar{x})$, therefore directly instructing the $LM_\phi$ learning objective (i.e., filling the $[mask]$), connecting text $\bar{x}$, disease $d_j$, and three aspects of disease $f$; (2) the concepts $o_{j_j}$ of ontology $O_j$ is injected into the verbalizer $V_{d_j}$, which projects the original prediction goals (i.e., $\bar{x}$ is a "symptom," "life event," or "treatment" of a mental disorder $d_j$) to a set of label words (i.e., $o_{j_j}$). As the prediction goal is a binary classification problem, the verbalizer words for negative examples are designed using manual verbalization methods, incorporating the most frequent words with the highest sentiment tendency.

Take "depression" as an example,

$$\begin{aligned} T_{depression}(\bar{x}) &= \textit{"}\bar{x}\textit{ is a [mask]}_{depression}\textit{ of depression symptom; }\bar{x}\textit{ is a [mask]}_{life_event}\textit{ of} \\ &\quad\ \textit{depression life event; }\bar{x}\textit{ is a [mask]}_{treatment}\textit{ of depression treatment."} \\ V_{depression}[mask]_{symptom} &= \textit{\{anxiety, dejected mood, ...\},} \\ V_{depression}[mask]_{life_event} &= \textit{\{divorce, domestic violence, ...\},} \\ V_{depression}[mask]_{treatment} &= \textit{\{supportive psychotherapy, abilify,...\}.} \end{aligned}$$

Essentially, $f_{prompt_rule}(\bar{x})$ instructs the $LM_\phi$ to evaluate whether $\bar{x}$ discloses disease $d_j$'s symptoms, life events, or treatments. This task is much simpler compared to the original user-level mental detection task and, therefore, aids the $LM_\phi$ in performance improvement. For instance, consider the case where $\bar{x}$ = "$I\ feel\ so\ lost\ after\ my\ divorce$ ....". The $f_{prompt_rule}(\bar{x})$ directs the $LM_{\phi}$ to discern whether $\bar{x}$ represents a symptom, a major life event, or a treatment associated with depression. The feedback from the $LM_{\phi}$ yields probabilities: $p(y^{symptom}_{depression}|\ \bar{x})\ =\ 0.5996$, $p(y^{life_event}_{depression}|\ \bar{x})\ =\ 0.8789$, and $p(y^{treatment}_{depression}|\ \bar{x})\ =\ 0.0001$.

This approach significantly reduces the difficulty for the $LM_{\phi}$ in determining whether a given $\bar{x}$ is related to $d_j$'s symptoms, life events, or treatments, allowing the $LM_{\phi}$ to focus on key areas already summarized by existing medical knowledge as the verbalizers are derived from disease ontology, helping the $LM_{\phi}$ transform inputs into prediction labels (Liu et al. 2023).

### 3.2.3. Prompt Ensembling of Multi-prompt Engineering for Mental Disorder Detection

The prompt engineering methods, $f_{prompt_prefix}(\cdot)$ and $f_{prompt_rule}(\cdot)$, focused on constructing a single prompt for different motivations for the mental disorder detection task using user-generated textual content. We now employ the prompt ensemble method to generate our multi-prompt function, $f_{prompt}(\cdot)$, for two reasons: (1) both $f_{prompt_prefix}(\cdot)$ and $f_{prompt_rule}(\cdot)$ are crucial in the context of mental disorder detection, and we need to combine them to accomplish the task together, (2) a significant body of research has demonstrated that the use of multiple prompts can further improve the efficacy of prompting methods (Liu et al. 2023).

While there are several methods for creating a multi-prompt function, we have opted for the prompt ensemble method, which involves utilizing multiple prompts for a given input during the inference phase to make predictions. It serves three purposes: (1) leveraging the complementary advantages of both $f_{prompt_prefix}(\cdot)$ and $f_{prompt_rule}(\cdot)$, (2) addressing the challenges of prompt engineering by eliminating the need to select a single best-performing prompt, and (3) stabilizing performance on downstream tasks (Liu et al. 2023). Formally,

$$f_{prompt}(\bar{x}) = \begin{cases} T_{d_j}(\bar{x}) = [v \oplus T_{d_j}^{symptom}(\bar{x}); T_{d_j}^{life_event}(\bar{x}); T_{d_j}^{treatment}(\bar{x})] \\ V_{d_j} = \{V_{d_j}[mask]_{symptom}, V_{d_j}[mask]_{life_event}, V_{d_j}[mask]_{treatment}\} \end{cases} \tag{12}$$

Note: if the $\bar{x} = \{t_1, t_2, ..., t_{|\bar{x}|}\}$ originates from the same $x = \{t_1, t_2, ..., t_{|x|}\}$, these $\bar{x}$ share the same prefix vector $v$.

The input to the $LM_\phi$ is the numerical vector representation of $T_{d_j}(\bar{x})$ which depends on the tokenizing method of $LM_\phi$, namely, $[v \oplus LM_tokenizing(T_{d_j}^{symptom}(\bar{x}); T_{d_j}^{life_event}(\bar{x}); T_{d_j}^{treatment}(\bar{x}))]$. The expected feedback from the model is

$$z_i = \begin{cases} P_\theta[:k], & \text{if } i \le k, \\ [mask]_i, z_{<i}, & \text{o.w.} \end{cases} \tag{13}$$

The information within the $[mask]$ is contingent upon two key factors: (1) $P_\theta$, serving as the prefix context, where all subsequent feedback hinges on the activations from the preceding feedback; (2) the template $T_{d_j}(\bar{x})$, which provides direct instructions and context to the $LM_\phi$; constraint how $LM_\phi$ fills in the content of $[mask]$ in $T_{d_j}(\bar{x})$.

The learning goal of our multi-prompt learning method is:

$$p(y_{d_j}|x) = \frac{1}{\lambda m} \sum_{f=1}^{r} \sum_{i=1}^{m} p_\phi\left([mask]_f = LM_\phi\left(y_{d_j}\right) | T_{d_j}(\bar{x})\right) \tag{14}$$

where $r$ is the number of masked positions in $f_{prompt}(\bar{x})$ (in our context $r = 3$), $[mask]_f = LM_\phi\left(y_{d_j}\right)$ is to map the class $y_{d_j}$ to the set of label words $V_{d_j}[mask]_f$, and $m$ is the number of sliding windows $\bar{x}$ in $x$.

The normalization term $\frac{1}{\lambda m}$ is introduced in Eq. 14 for two reasons: (1) $p(\cdot)$ represents probability feedback from $LM_\phi$, and the sum of multiple probabilities could exceed 1. The upper limit of $\sum_{j=1}^{r} \sum_{i=1}^{m} p(\cdot)$ is $m + r$ (if each $p(\cdot)$ returns a value of 1). Since $r$ is a very small number in our setting, we simplify the upper limit to $m$; the lower limit of this summation of multiple probabilities is 0 (if each $p(\cdot)$ returns a value of 0). Consequently, if we normalize $\sum_{i=1}^{m} p(\cdot)$ to the range $[0, 1]$, $(\sum_{i=1}^{m} p(\cdot) - lower_limit)/(upper_limit - lower_limit)(1 - 0) + 0 = \frac{1}{m}\sum_{i=1}^{m} p(\cdot)$. (2) Additionally, since we employ a sliding window $\bar{x}$ to break down $x$ and thereby simplify the mental disorder detection task, a possibility arises: multiple sliding window $\bar{x}$ instances may be describing the same symptoms, life events, or treatment for a disease $d_j$. Therefore, we also use $\frac{1}{m}$ as a penalization factor. Overall, we incorporate the normalization term $\frac{1}{\lambda m}$ into our learning goal, where λ is a hyperparameter.

# 4. EVALUATION

To evaluate the performance of the proposed method, we conduct the following examinations: (1) Comparison with benchmarks: we demonstrate the advantages of LLM-based prompt engineering over other machine learning paradigms in our research context. (2) Comparison with other prompt strategies: we highlight the benefits of continuous prompts over discrete prompts in maximizing the capabilities of LLMs within our binary mental disorder detection/classification task. (3) Ablation studies: we analyze the contribution of each component in our ensemble prompt to overall performance. (4) We present three experiments to illustrate the unique advantages of prompt engineering: few-shot learning, early identification, and generalizability to scenarios with very limited labeled data. (5) Post-analysis to explain the effectiveness of our method.

## 4.1. Experimental Setup

All prompt engineering methods rely on pre-trained LLMs to accomplish downstream tasks. Therefore, the major characteristics of pre-trained LLMs, including their main training objective, type of text noising, auxiliary training objective, attention mask, and typical architecture, can significantly influence the performance of downstream tasks in the pre-train and prompt engineering learning paradigm (Liu et al. 2023). Thus, we review mainstream LLM architectures and describe how we chose one for our experiments. Most widely used LLMs adopt the Transformer architecture, which can be classified into three types based on their characteristics (see Table 3): decoder models, encoder models, and encoder-decoder models.

**Table 3. Summary of Mainstream Transformer-based LLMs**

| Category | Training | Examples |
|---|---|---|
| Decoder models (*Auto-regressive models*) | • Use only the decoder of a Transformer model.<br>• At each stage, for a given word the attention layers can only access the words positioned before it in the sentence.<br>• Pretraining: predicting the next word in the sentence. | CTRL, GPT, GPT-2, Transformer XL, DeepSeek-R1 |
| Encoder models (*auto-encoding models*) | • Use only the encoder of a Transformer model.<br>• At each stage, the attention layers can access all the words in the initial sentence (i.e., bi-directional attention).<br>• Pretraining: corrupting a given sentence (e.g., masking random words) and tasking the model with finding or reconstructing the initial sentence. | ALBERT, BERT, DistilBERT, ELECTRA, RoBERTa |
| Encoder-decoder models (*sequence-to-sequence models*) | • Use both parts of the Transformer architecture.<br>• At each stage, the encoder attention layers can access all the words in the initial sentence, whereas the decoder attention layers can only access the<br>• words positioned before a given word in the input.<br>• Pretraining: using the objectives of encoder or decoder models, or replacing random words and predicting the masked words. | BART, Marian, T5 |

Given the specific nature of our experimental environment[2], prompt design (continuous prompt, prefix tuning, and rule-based prompt), prediction task (binary classification for subject-level mental disorder detection), and research context (chronic disease predictive analytics), we employed an exclusion-based approach to select appropriate LLMs. (1) We exclude encoder models (e.g., BERT and RoBERTa) since our prompt design incorporates prefix-tuning to achieve personalized prompts. The implementation of prefix tuning requires LLMs to allow the injection of past_key_values of HuggingFace LLMs (caching and reusing the intermediate hidden states from the previous steps) into the model. Encoder models on HuggingFace do not permit such operations. (2) We exclude decoder models because previous studies show that for supervised learning tasks with input-label pair data (as opposed to self-supervised learning), encoder-decoder models offer advantages by requiring fewer parameters and achieving better results compared to decoder-only models (Jiang et al. 2023). (3) We exclude non-open-source LLMs (e.g., GPT-4o) since our approach relies on continuous prompt engineering, which requires open-source LLMs to operate in the embedding space (However, non-open-source LLMs with discrete prompt can still be used as benchmark models). Moreover, since healthcare analytics problems often involve user-generated data or fine-grained patient-level information, utilizing such models could pose potential privacy issues. As a result, we selected FLAN-T5 (one of the best-performing open-source encoder-decoder models) as the base LLM for our experiments.

[2] We used pre-trained LLMs from Hugging Face (https://huggingface.co/), OpenPrompt for prompt implementation (https://github.com/thunlp/OpenPrompt), PyTorch ver. 2.0.1 with GPU: GeForce RTX 4090.

For evaluation, we mainly use three datasets (Table 4) from the eRisk database (Losada et al. 2018, Parapar et al. 2021). Specifically, we selected the detection of *depression*, *anorexia*, and *pathological gambling* as the primary tasks given their prevalence[3] and broad societal impact.

To assess the usability of the methods to new users, we use 60% of the data for training, 20% for validation, and 20% for testing. The reported results are the average performances of 10 experiment runs. We also report the standard deviation of these experiments to show our results' statistical significance. In our evaluations, we report AUC, F1-score, precision, and recall. The goal of the proposed method is to achieve the highest AUC and F1-score. It is important to note that in our research context, a model with high precision but low recall, or vice versa, does not necessarily indicate good overall performance. When precision (exactness: correctly identifying true patients) is low, false-positive patients will suffer from unnecessary mental burden and diagnostic costs. When recall (completeness: capturing as many patients as possible) is low, the practical utility of the model is compromised.

**Table 4. Datasets Summary**

| Dataset | | # of subjects | # of posts | # of words | Avg # of posts per subject | Avg # of days from first to last post |
|---|---|---|---|---|---|---|
| Depression | *P* | 214 | 90,222 | 2,480,216 | 421 | 676 |
| | *N* | 1,493 | 986,360 | 22,461,242 | 660 | 664 |
| Anorexia | *P* | 134 | 42,493 | 1,583,227 | 317 | 679 |
| | *N* | 1,153 | 781,768 | 16,781,263 | 678 | 848 |
| Pathological gambling | *P* | 245 | 69,301 | 2,119,872 | 282 | 545 |
| | *N* | 4,182 | 2,088,002 | 44,077,018 | 499 | 663 |

*Note:* P = positive examples; N = negative examples.

### 4.2. Comparison with Benchmark Methods

#### 4.2.1. Comparison with Existing Mental Disorder Detection Methods

[3] According to the World Health Organization (WHO), approximately 306,356,000 people worldwide struggle with depression, 14,000,000 individuals suffer from eating disorders, and between 8,062,000 and 483,720,000 people worldwide struggle with pathological gambling.

We begin by comparing our proposed method with state-of-the-art approaches from other machine learning paradigms other than prompt engineering. The goal is to highlight the advantages of LLM-based prompt engineering over these alternative paradigms in the context of our research, which focuses on detecting mental disorders through user-generated content. Table 5 reports our results and compares with state-of-the-art methods for benchmarking (Benton et al. 2017, Chau et al. 2020, Choudhury et al. 2013, Coppersmith et al. 2014, Khan et al. 2021, Lin et al. 2020, Malviya et al. 2021, Preoţiuc-Pietro et al. 2015, Reece et al. 2017).

**Table 5. Comparison with State-of-the-art Mental Disorder Detection Studies**

| Dataset | Model | | AUC | F1 | Precision | Recall |
|---|---|---|---|---|---|---|
| Depression | Traditional Machine Learning with Feature Engineering | Choudhury et al. (2013) | 0.569 ± 0.001 | 0.588 ± 0.002 | 0.716 ± 0.001 | 0.569 ± 0.003 |
| | | Coppersmith et al. (2014) | 0.685 ± 0.002 | 0.705 ± 0.001 | 0.735 ± 0.003 | 0.685 ± 0.002 |
| | | Preoţiuc-Pietro et al. (2015) | 0.723 ± 0.003 | 0.760 ± 0.002 | 0.820 ± 0.001 | 0.720 ± 0.001 |
| | | Benton et al. (2017) | 0.716 ± 0.029 | 0.722 ± 0.026 | 0.730 ± 0.022 | 0.716 ± 0.029 |
| | | Reece et al. (2017) | 0.729 ± 0.012 | 0.717 ± 0.011 | 0.708 ± 0.013 | 0.729 ± 0.011 |
| | | Chau et al. (2020) | 0.623 ± 0.006 | 0.573 ± 0.005 | 0.570 ± 0.002 | 0.622 ± 0.005 |
| | Deep Learning with Representati on Learning | CNN-based (Lin et al. 2020) | 0.710 ± 0.005 | 0.711 ± 0.004 | 0.728 ± 0.018 | 0.710 ± 0.005 |
| | | LSTM-based (Khan et al. 2021) | 0.765 ± 0.004 | 0.756 ± 0.004 | 0.751 ± 0.008 | 0.765 ± 0.004 |
| | | Transformer-based (Malviya et al. 2021) | 0.751 ± 0.002 | 0.734 ± 0.005 | 0.724 ± 0.010 | 0.751 ± 0.002 |
| | **Ours** | Flan-T5 + Prefix Tuning + Rule based prompt | **0.915 ± 0.006** | **0.913 ± 0.003** | 0.888 ± 0.005 | 0.939 ± 0.005 |
| Anorexia | Traditional Machine Learning with Feature Engineering | Choudhury et al. (2013) | 0.669 ± 0.007 | 0.708 ± 0.009 | 0.793 ± 0.005 | 0.669 ± 0.007 |
| | | Coppersmith et al. (2014) | 0.800 ± 0.001 | 0.812 ± 0.003 | 0.826 ± 0.001 | 0.800 ± 0.001 |
| | | Preoţiuc-Pietro et al. (2015) | 0.748 ± 0.006 | 0.758 ± 0.007 | 0.770 ± 0.002 | 0.748 ± 0.006 |
| | | Benton et al. (2017) | 0.714 ± 0.005 | 0.759 ± 0.007 | 0.877 ± 0.007 | 0.714 ± 0.000 |
| | | Reece et al. (2017) | 0.704 ± 0.005 | 0.770 ± 0.007 | 0.963 ± 0.007 | 0.704 ± 0.005 |
| | | Chau et al. (2020) | 0.707 ± 0.009 | 0.733 ± 0.002 | 0.770 ± 0.005 | 0.707 ± 0.009 |
| | Deep Learning with Representati on Learning | CNN-based (Lin et al. 2020) | 0.481 ± 0.009 | 0.601 ± 0.009 | 0.488 ± 0.007 | 0.783 ± 0.001 |
| | | LSTM-based (Khan et al. 2021) | 0.500 ± 0.000 | 0.666 ± 0.007 | 0.500± 0.000 | 1.000 ± 0.000 |
| | | Transformer-based (Malviya et al. 2021) | 0.771 ± 0.001 | 0.743 ± 0.002 | 0.846 ± 0.002 | 0.662 ± 0.007 |
| | **Ours** | Flan-T5 + Prefix Tuning + Rule based prompt | **0.886 ± 0.005** | **0.877 ± 0.006** | 0.824 ± 0.003 | 0.938 ± 0.004 |
| Pathological gambling | Traditional Machine Learning with Feature Engineering | Choudhury et al. (2013) | 0.637 ± 0.005 | 0.704 ± 0.012 | 0.637 ± 0.001 | 0.976 ± 0.003 |
| | | Coppersmith et al. (2014) | 0.534 ± 0.006 | 0.549 ± 0.003 | 0.534 ± 0.002 | 0.970 ± 0.002 |
| | | Preoţiuc-Pietro et al. (2015) | 0.586 ± 0.005 | 0.633 ± 0.003 | 0.586 ± 0.005 | 0.978 ± 0.004 |
| | | Benton et al. (2017) | 0.575 ± 0.109 | 0.547 ± 0.066 | 0.575 ± 0.109 | 0.585 ± 0.119 |
| | | Reece et al. (2017) | 0.551 ± 0.001 | 0.573 ± 0.005 | 0.551 ± 0.008 | 0.871 ± 0.001 |
| | | Chau et al. (2020) | 0.917 ± 0.034 | 0.772 ± 0.048 | 0.917 ± 0.034 | 0.717 ± 0.040 |
| | Deep Learning with Representati on Learning | CNN-based (Lin et al. 2020) | 0.818 ± 0.002 | 0.779 ± 0.006 | 0.989 ± 0.005 | 0.643 ± 0.002 |
| | | LSTM-based (Khan et al. 2021) | 0.610 ± 0.002 | 0.713 ± 0.009 | 0.563 ± 0.009 | 0.972 ± 0.007 |
| | | Transformer-based (Malviya et al. 2021) | 0.687 ± 0.005 | 0.551 ± 0.004 | 0.976 ± 0.009 | 0.384 ± 0.001 |
| | **Ours** | Flan-T5 + Prefix Tuning + Rule based prompt | **0.989 ± 0.000** | **0.985 ± 0.002** | 0.985 ± 0.002 | 0.990 ± 0.003 |

Existing mental disorder detection methods that utilize user-generated content can be mainly classified into two categories: (1) traditional machine learning methods with feature engineering and (2) deep neural networks with representation learning. Our proposed method shows significant performance improvement compared to models in both categories. Compared with the best-performing model in traditional machine learning, our method improves the F1 score by 0.153 on the depression dataset (Preoţiuc-Pietro et al., 2015), 0.065 on the anorexia dataset (Coppersmith et al., 2014), and 0.213 on the pathological gambling dataset (Chau et al. 2020). Compared with the best-performing deep learning model, our method improves the F1 score by 0.157 on the depression dataset (Khan et al. 2021), 0.134 on the anorexia dataset (Malviya et al. 2021), and 0.206 on the pathological gambling dataset (Lin et al. 2020).

Our method consistently outperforms state-of-the-art mental disorder detection methods across various disorders and datasets for the following reasons: (1) Noisy data: Capturing valid information from noisy user-generated content is a significant challenge for traditional machine learning models that rely on feature engineering. Similarly, conventional deep learning models may struggle to memorize text sequences (e.g., LSTM) that are exceptionally lengthy and exhibit a wide range of lengths (e.g., Transformers). In contrast, our method using LLMs with prompt engineering is capable of extracting task-specific information from noisy text data. (2) Class imbalance: Moreover, the dataset is highly imbalanced (around six times more negative samples compared to positive samples), which reflects the real-world distribution of depression patients. It can be highly challenging for the baseline models to excel in this context, while our method can resolve this issue with personalized prompts and knowledge injection.

However, we acknowledge that standard machine learning and deep learning methods are trained using only the given dataset. Moreover, they have relatively fewer model parameters compared to the LLMs used for the experiment (i.e., Flan-T5). Thus, the significant performance gain could be partially due to the superior capability of the LLMs themselves. Nonetheless, *we would like to clarify that the inherent ability of LLMs is not our main contribution. Our contribution focuses on prompt engineering design, which can further enhance the capability of*

*LLMs*. To support this, we will demonstrate the effectiveness of multi-prompt engineering by comparing it with other prompt strategies (Section 4.2.2) and presenting the results of ablation studies (Section 4.3).

#### 4.2.2. Comparison with Other Prompt Engineering Strategies

To further validate the effectiveness of our prompt engineering method, we compare it with other prompt engineering strategies (Table 6). (1) Continuous prompting vs. discrete prompting: Our method employs a continuous prompting strategy to achieve optimal classification results. In contrast, another approach is to directly query the LLM using a discrete prompt (i.e., human-understandable language) for mental disorder detection. (2) Prefix tuning in continuous prompting vs. personalized context in discrete prompting: Within our ensemble prompt design, we incorporate prefix tuning to capture personalized context among social media users and patients. A similar motivation in discrete prompting is context tailored to the social media user (i.e.,using a small set of labeled cases as examples to provide context before the LLM generates feedback). We compare the effects of these two approaches. (3) Rule-based prompting in continuous prompting vs. chain-of-thought reasoning in discrete prompting: Our ensemble prompt design includes rule-based prompting to break down the complex mental disorder detection problem into subtasks including identifying symptoms, life events, and treatments. In discrete prompting, a similar motivation is achieved through chain-of-thought reasoning (i.e., prompting an LLM with examples that break down a task into several sub-components can enhance model performance) (Wei et al. 2022). Following this approach, we provided brief reasoning as a chain of thoughts within a discrete prompt, guided by high-quality examples of mental disorder diagnoses that include relevant symptoms, major life events, and treatment. Building on (2), we further compare the effects of these two approaches.

According to our results, we observed several important findings: When comparing the performance of continuous and discrete prompts using the same LLM, continuous prompts significantly outperformed discrete prompts across all three datasets in terms of F1 score (0.611 vs. 0.529 for depression, 0.727 vs. 0.551 for anorexia, and 0.705 vs. 0.600 for pathological gambling). This improvement can be attributed to the fundamental difference between the two approaches: discrete prompts rely on predefined, natural language-based templates to achieve the prediction goal, whereas continuous prompts leverage trainable parameters to adjust prompt embeddings (Liu et al. 2023). In our task, optimizing predictive performance is of primary importance. Because continuous prompts can adjust their embeddings through trainable parameters, they have a greater potential to minimize loss.

**Table 6. Comparison with Other Prompt Engineering Strategies**

| Dataset | Prompt Strategies | | AUC | F1 | Precision | Recall |
|---|---|---|---|---|---|---|
| Depression | *Continuous* | Flan-T5 | **0.673 ± 0.005** | **0.611 ± 0.002** | 0.578 ± 0.002 | 0.647 ± 0.006 |
| | *Discrete prompt* | Flan-T5 | **0.673 ± 0.002** | 0.529 ± 0.001 | 0.368 ± 0.009 | 0.940 ± 0.001 |
| Anorexia | *Continuous* | Flan-T5 | **0.783 ± 0.000** | **0.727 ± 0.003** | 0.578 ± 0.003 | 0.979 ± 0.006 |
| | *Discrete prompt* | Flan-T5 | 0.623 ± 0.007 | 0.551 ± 0.007 | 0.524 ± 0.006 | 0.581 ± 0.008 |
| Pathological gambling | *Continuous* | Flan-T5 | **0.772 ± 0.007** | **0.705 ± 0.009** | 0.545 ± 0.005 | 1.000 ± 0.000 |
| | *Discrete prompt* | Flan-T5 | 0.746 ± 0.003 | 0.600 ± 0.008 | 0.908 ± 0.005 | 0.448 ± 0.008 |
| Depression | *Continuous* | Flan-T5 + Prefix Tuning | **0.704 ± 0.007** | **0.625 ± 0.003** | 0.488 ± 0.004 | 0.869 ± 0.000 |
| | *Discrete prompt* | Flan-T5 + Personalized context | 0.601 ± 0.000 | 0.368 ± 0.009 | 0.236 ± 0.008 | 0.833 ± 0.003 |
| Anorexia | *Continuous* | Flan-T5 + Prefix Tuning | **0.787 ± 0.005** | **0.784 ± 0.008** | 0.775 ± 0.000 | 0.794 ± 0.009 |
| | *Discrete prompt* | Flan-T5 + Personalized context | 0.474 ± 0.005 | 0.310 ± 0.007 | 0.262 ± 0.003 | 0.381 ± 0.000 |
| Pathological gambling | *Continuous* | Flan-T5 + Prefix Tuning | **0.881 ± 0.008** | **0.868 ± 0.007** | 0.781 ± 0.008 | 0.977 ± 0.003 |
| | *Discrete prompt* | Flan-T5 + Personalized context | 0.722 ± 0.005 | 0.590 ± 0.002 | 0.658 ± 0.005 | 0.534 ± 0.007 |
| Depression | *Continuous* | Flan-T5 + Prefix Tuning + Rule based prompt **(Ours)** | **0.915 ± 0.000** | **0.913 ± 0.000** | 0.888 ± 0.000 | 0.939 ± 0.000 |
| | *Discrete prompt* | Flan-T5 + Personalized context + Chain of thought | 0.614 ± 0.001 | 0.401 ± 0.006 | 0.263 ± 0.002 | 0.847 ± 0.005 |
| Anorexia | *Continuous* | Flan-T5 + Prefix Tuning + Rule based prompt **(Ours)** | **0.886 ± 0.005** | **0.877 ± 0.006** | 0.824 ± 0.003 | 0.938 ± 0.004 |
| | *Discrete prompt* | Flan-T5 + Personalized context + Chain of thought | 0.560 ± 0.003 | 0.454 ± 0.005 | 0.409 ± 0.008 | 0.510 ± 0.002 |
| Pathological gambling | *Continuous* | Flan-T5 + Prefix Tuning + Rule based prompt **(Ours)** | **0.989 ± 0.000** | **0.985 ± 0.002** | 0.985 ± 0.002 | 0.990 ± 0.003 |
| | *Discrete prompt* | Flan-T5 + Personalized context + Chain of thought | 0.634 ± 0.006 | 0.489 ± 0.001 | 0.682 ± 0.009 | 0.381 ± 0.000 |

Note.

- Discrete prompt: "*Below is a collection of messages from a single user on a social media platform. Based on the content of these messages, can it be determined whether this individual is experiencing {depression, anorexia, pathological gambling}? <insert posts from an individual user>"*
- Discrete prompt + Personalized context: *Researchers utilize social media to detect whether individuals exhibit signs of depression. Below are examples:*

*Example 1: " <insert one post from a single user>." This individual is identified as having {depression, anorexia, pathological gambling}.*

*…..*

*Below is a collection of messages from a single user on a social media platform. Based on the content of these messages, can it be determined whether this individual is experiencing {depression, anorexia, pathological gambling}?*

*Messages: "<insert the message from an individual user>"*

- Discrete prompt + Personalized context + Chain of thought: *Researchers utilize social media to detect whether individuals exhibit signs of {depression, anorexia, pathological gambling}. Below are examples:*

*Example 1: "<insert one post from a single user>." This individual exhibits symptoms of {depression, anorexia, pathological gambling}, as evidenced by {depression, anorexia, pathological gambling} symptoms such as "I just want to die," "everything hurts so much," and "morning nausea." Consequently, this individual is identified as having {depression, anorexia, pathological gambling}.*

*Below is a collection of messages from a single user on a social media platform. Based on the content of these messages, can it be determined whether this individual is experiencing {depression, anorexia, pathological gambling}?*

*Messages: "<insert the message from an individual user>"*

Next, we compare two different prompt strategies for providing personalized context: prefix tuning in continuous prompts and directly incorporating personalized context in discrete prompts. As shown in our results, prefix tuning in continuous prompts consistently outperformed directly incorporating personalized context in discrete prompts across all three datasets in terms of F1 score (0.625 vs. 0.368 for depression, 0.784 vs. 0.310 for anorexia, and 0.868 vs. 0.590 for pathological gambling). We argue that the performance improvements observed in continuous prompts can be attributed to several key factors. First, prefix tuning as a continuous prompt method leverages trainable parameters to adjust prompt embeddings, allowing each individual's personalized context to be effectively captured; additionally, the relationships between individuals are also embedded through the tuning of these trainable parameters, ultimately enhancing the prediction outcomes (Li and Liang 2021). In contrast, discrete prompts rely on pre-established templates to achieve the prediction goal. In our research context, a complex mental disorder detection task, the patient's context is incorporated into the prompt alongside the prediction task. This approach significantly increases the difficulty of comprehending the task for the LLM. Moreover, information cannot be shared across individuals, since LLMs do not retain memory of previous cases, further limiting predictive performance. Our experimental results show that adding personalized context to discrete prompts not only fails to improve prediction accuracy but actually degrades performance.

Next, we compare two different prompt strategies for incorporating domain knowledge: rule-based prompts in continuous prompting and chain-of-thought prompting in discrete prompting. As shown in our results, rule-based prompts in continuous prompting significantly outperformed chain-of-thought prompting in discrete prompting across all three datasets in terms of F1 score (0.913 vs. 0.401 for depression, 0.877 vs. 0.454 for anorexia, and 0.985 vs. 0.489 for pathological gambling). We argue that several factors contribute to this performance improvement. First, the essence of rule-based prompts lies in leveraging domain knowledge to break down complex problems into smaller, more manageable ones (Han et al. 2022). In our research design, these subproblems, informed by a structured mental disorder ontology from medical knowledge, lead to clearer prediction targets, a more comprehensive understanding of medical concepts, and more efficient use of domain knowledge. Collectively, this enhances the model's predictive capabilities. In contrast, chain-of-thought prompting provides reasoning steps and examples to the LLM (e.g., "This patient is diagnosed with depression because they experienced life event B and exhibit symptom A…"). However, this method introduces medical knowledge in a fragmented and incomplete manner, offering LLMs only limited assistance from domain knowledge. As a result, its contribution to performance improvement remains minimal.

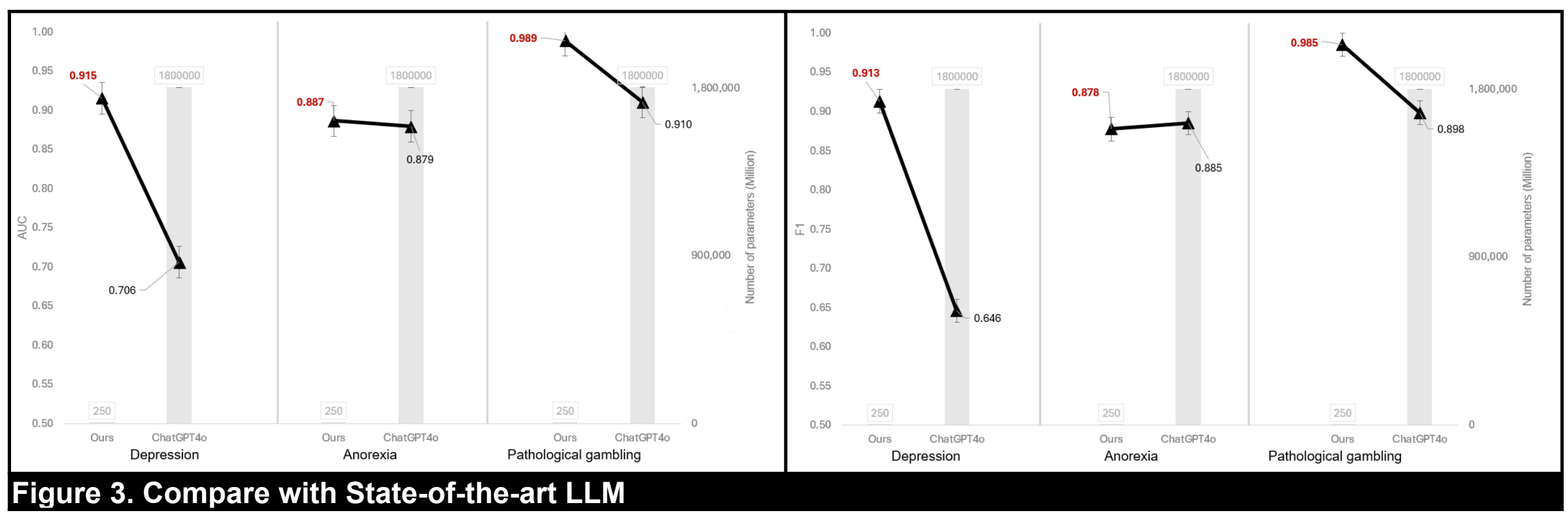

**Figure 3. Compare with State-of-the-art LLM**

*Note:* **Ours**: (1) Continuous prompt: Prefix Tuning + Rule-based prompt; (2) Number of parameters: Flan-T5 (Base): 250 Million
**ChatGPT4o**: (1) Discrete prompt: Personalized context + Chain of thought; (2) Number of parameters: ChatGPT4o: 1800,000 Million

To further highlight the advantages of our proposed continuous prompt-based strategy over discrete prompting, we compare our method using FLAN-T5 (Continuous Prompt: Prefix Tuning + Rule-based Prompt) with a discrete prompt approach using ChatGPT-4o (Discrete Prompt: Personalized Context + Chain of Thought). The difference in the number of parameters between these two LLMs is substantial: FLAN-T5 has 250 million parameters, whereas ChatGPT-4o has 1.8 trillion. Generally speaking, the number of parameters reflects an LLM's baseline capability when no additional techniques are applied. Despite the significant disparity in model size, our approach outperforms ChatGPT-4o with discrete prompting across three datasets in most cases in terms of F1 score (e.g., 0.913 vs. 0.646 for depression and 0.985 vs. 0.898 for pathological gambling). On the anorexia dataset, the performance of our method is comparable to that of ChatGPT-4o with discrete prompting (0.877 vs. 0.885).

The experiments described above provide several insights. Our research focuses on subject-level user-generated content-based mental disorder detection, which is a binary classification problem. Given this context, continuous prompts can fully leverage the capabilities of LLMs to achieve optimal classification and prediction performance. Additionally, user-generated content contains significant noise, and there is substantial individual variability among patients. Given this context, effectively incorporating existing medical knowledge and highlighting individual differences can enhance the performance of LLMs. Continuous prompts have advantages over discrete prompts in both aspects: In our research design, prefix tuning enables the effective capture of personalized context, and it also embeds relationships between individuals through parameter tuning. Moreover, our rule-based tuning approach systematically integrates medical knowledge, further improving the model's performance. Furthermore, unlike discrete prompts, which increase the model's workload when incorporating personalized context and chains of thought, our approach alleviates the burden of LLMs in understanding the prediction task, thereby improving predictive performance.

In the experiments above, we observed that continuous prompts demonstrate significant performance improvements over discrete prompts. At the same time, they are computationally efficient. Discrete prompts may initially appear "low-cost" in computational terms due to the absence of training requirements; however, optimizing them for performance demands substantial time investments. In contrast, continuous prompts automate this optimization process with negligible computational overhead. Compared to full fine-tuning of LLMs, continuous prompts provide a streamlined alternative: they involve training only small, parameterized vectors appended to the input embeddings, reducing the training time to approximately 10% of that required for full LLM fine-tuning. This efficiency-performance trade-off is consistent with patterns documented in prior research on continuous prompt methods (Lester et al. 2021, Liu et al. 2024).

### 4.3. Ablation Studies

Since our prompt engineering approach employs an ensemble method with multiple prompt components, we conduct ablation studies to assess the relative impact of each component on mental disorder detection tasks.

We first assess the impact of our ensemble prompt engineering method compared to just using an LLM for prediction (Table 7). With its more sophisticated structure and a large number of parameters, Flan-T5 outperforms most baseline methods as we expected (see Table 6). However, adding our prompt engineering method further improves the predictive performance by a significant margin in terms of F1 score (i.e., 0.913 vs. 0.529 for depression, 0.877 vs. 0.724 for anorexia, and 0.985 vs. 0.705 for pathological gambling). This result ensures that the performance gain of our method is not just achieved by the predictive capabilities of LLMs. The innovative aspects of the proposed prompting solution further enhance the predictive performance.

**Table 7. Ablation Studies**

| Dataset | Prompt Strategies | | AUC | F1 | Recall | Precision |
|---|---|---|---|---|---|---|
| Depression | **Ours** - Prefix tuning - Rule based prompt | Flan-T5 | 0.672 ± 0.005 | 0.529 ± 0.002 | 0.368 ± 0.002 | 0.940 ± 0.006 |
| | **Ours** - Rule based prompt | Flan-T5 + Prefix tuning | 0.704 ± 0.007 | 0.625 ± 0.003 | 0.488 ± 0.004 | 0.869 ± 0.000 |
| | **Ours** - Prefix tuning | Flan-T5 + Rule based prompt | 0.757 ± 0.008 | 0.794 ± 0.007 | 0.938 ± 0.000 | 0.689 ± 0.005 |
| | **Ours** | Flan-T5 + Prefix tuning + Rule based prompt | **0.915 ± 0.006** | **0.913 ± 0.003** | 0.888 ± 0.005 | 0.939 ± 0.005 |
| Anorexia | **Ours** - Prefix tuning - Rule based prompt | Flan-T5 | 0.783 ± 0.000 | 0.727 ± 0.003 | 0.578 ± 0.003 | 0.979 ± 0.006 |
| | **Ours** - Rule based prompt | Flan-T5 + Prefix tuning | 0.787 ± 0.005 | 0.784 ± 0.008 | 0.775 ± 0.000 | 0.794 ± 0.009 |
| | Ours - Prefix tuning | Flan-T5 + Rule based prompt | 0.850 ± 0.000 | 0.858 ± 0.008 | 0.912 ± 0.005 | 0.811 ± 0.001 |
| | **Ours** | Flan-T5 + Prefix tuning + Rule based prompt | **0.886 ± 0.005** | **0.877 ± 0.006** | 0.824 ± 0.003 | 0.938 ± 0.004 |
| Pathological gambling | **Ours** - Prefix tuning - Rule based prompt | Flan-T5 | 0.772 ± 0.007 | 0.705 ± 0.009 | 0.545 ± 0.005 | 1.000 ± 0.000 |
| | **Ours** - Rule based prompt | Flan-T5 + Prefix tuning | 0.881 ± 0.008 | 0.868 ± 0.007 | 0.781 ± 0.008 | 0.977 ± 0.003 |
| | **Ours** - Prefix tuning | Flan-T5 + Rule based prompt | 0.935 ± 0.002 | 0.932 ± 0.009 | 0.900 ± 0.000 | 0.968 ± 0.002 |
| | **Ours** | Flan-T5 + Prefix tuning + Rule based prompt | **0.989 ± 0.000** | **0.985 ± 0.002** | 0.985 ± 0.002 | 0.990 ± 0.003 |

Next, we assess the impact of removing the rule-based component from our proposed method. This removal significantly reduces prediction performance in terms of F1 score, as evidenced by the following results: depression (0.913 → 0.625), anorexia (0.877 → 0.784), and pathological gambling (0.985 → 0.868). Similarly, we evaluate the effect of removing the prefix-tuning component, which also leads to a substantial decline in performance: depression (0.913 → 0.794), anorexia (0.877 → 0.858), and pathological gambling (0.985 → 0.932). Overall, removing any components hinders the prediction performance, indicating the effectiveness of both components in our method.

### 4.4. The Advantage of LLM-based Prompt Engineering

In this section, we present three experiments that highlight the unique advantages of prompt engineering: few-shot learning, early identification, and generalizability to scenarios with limited labeled data. Due to page limits, we focus on the *depression* dataset for the few-shot learning and early identification tasks, as it is the most common mental disorder (WHO 2022b). Additionally, we include a *self-harm* dataset, which contains only 41 labeled positive cases, to demonstrate our method's ability to generalize to rare or emerging diseases with limited labeled data.

#### 4.4.1. Few-shot Learning Capability

One salient benefit of using prompt engineering compared to existing mental disorder detection methods is the ability to few-shot learn with minimal labeled data, powered by general-purpose language understanding capabilities. Thus, in this section, we assess the efficacy of our proposed prompt engineering method in accomplishing few-shot learning on the depression detection task. To this end, we substantially decreased the number of training examples from the original training set of 1,000 subjects (i.e., original training set $1,707 \times 60\%$) to 100, 10, and 2 subjects. Figure 4 shows the results of this experiment.

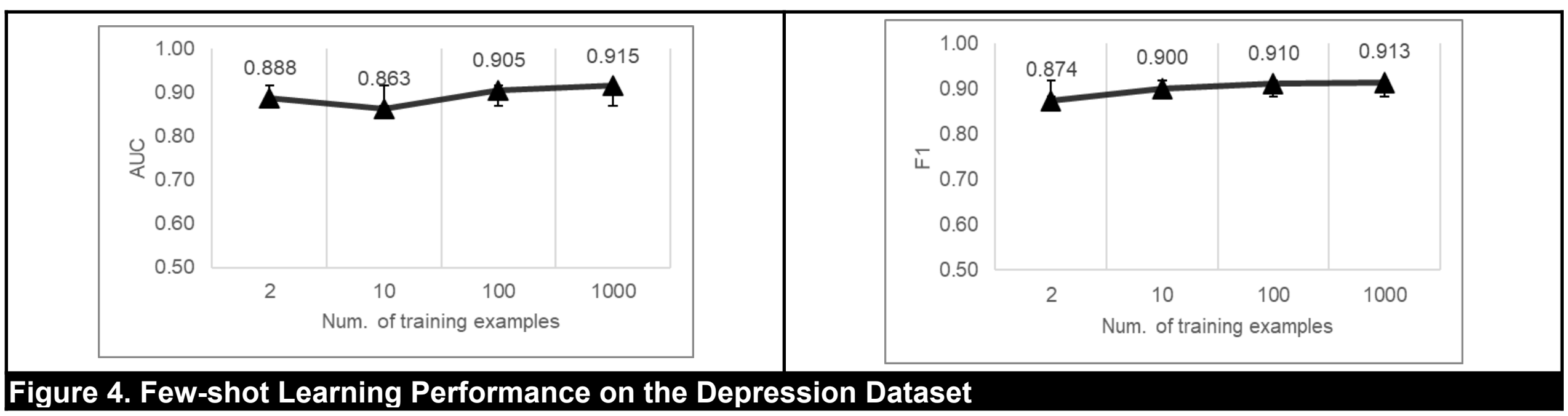

**Figure 4. Few-shot Learning Performance on the Depression Dataset**

The results indicate that, in contrast to the baseline models, our prompt engineering approach can achieve satisfactory performance even with a significantly smaller training set. In comparison to the best-performing benchmarks, our method demonstrates significant improvement in terms of F1 scores. For instance, the best-performing deep learning model and machine learning model on the depression dataset showed F1-scores of 0.756 and 0.760, respectively, while our method achieved 0.874 with just two training samples. Moreover, as the training size increases, there is a corresponding enhancement in terms of F1 score.

It is noteworthy that other models fail to converge or achieve satisfactory performance when the number of training examples is within the range of 2, 10, or 100. The superior performance of our prompt engineering method in few-shot learning has several important implications in the context of mental disorder detection. First, it reduces the demand for a large amount of labeled training data, significantly alleviating the burden on researchers and platforms and paving the way for large-scale, continuous monitoring of mental disorders. Furthermore, for emerging diseases with rare cases and scarce examples, our proposed method holds promise in quickly capturing the necessary predictive information through personalized prompts and medical knowledge injection with minimal training data, thus achieving satisfactory prediction outcomes (we also demonstrate the generalizability of our method to identifying rare mental disorders in a later section). It provides unique opportunities for flexible and rapid monitoring of the emergence and spread of new diseases in the population through user-generated content, which existing methods are unable to achieve.

#### 4.4.2. Early Identification of Depression

In addition to few-shot learning, early identification is essential when detecting mental disorders to provide timely warning cues to platforms, medical professionals, and potential patients. It facilitates timely reminders and proactive treatments, thereby preventing the exacerbation of mental disorders and the resulting burden of disease. Using depression as a research case, by taking the onset of depression as the anchor point, we incorporate data from various time frames before the onset of depression, specifically $x$ weeks before the onset of depression, where $x = \{2, 4, 8, 24\}$, while using a one-month (4 weeks) duration of data as input. A significant challenge in this early prediction lies in the fact that we only use a one-month duration of data as input, the average number of posts per user significantly decreases, and the information pertaining to depression becomes increasingly sparse (it may even lack relevant information associated with depression). The experiment results (see Figure 4) indicate that even in such a challenging context, our proposed prompt engineering method shows a relatively stable F1-score (ranging between 0.727 and 0.738). This result stands in comparison to traditional machine learning and deep learning models that rely on the entire available data for prediction.

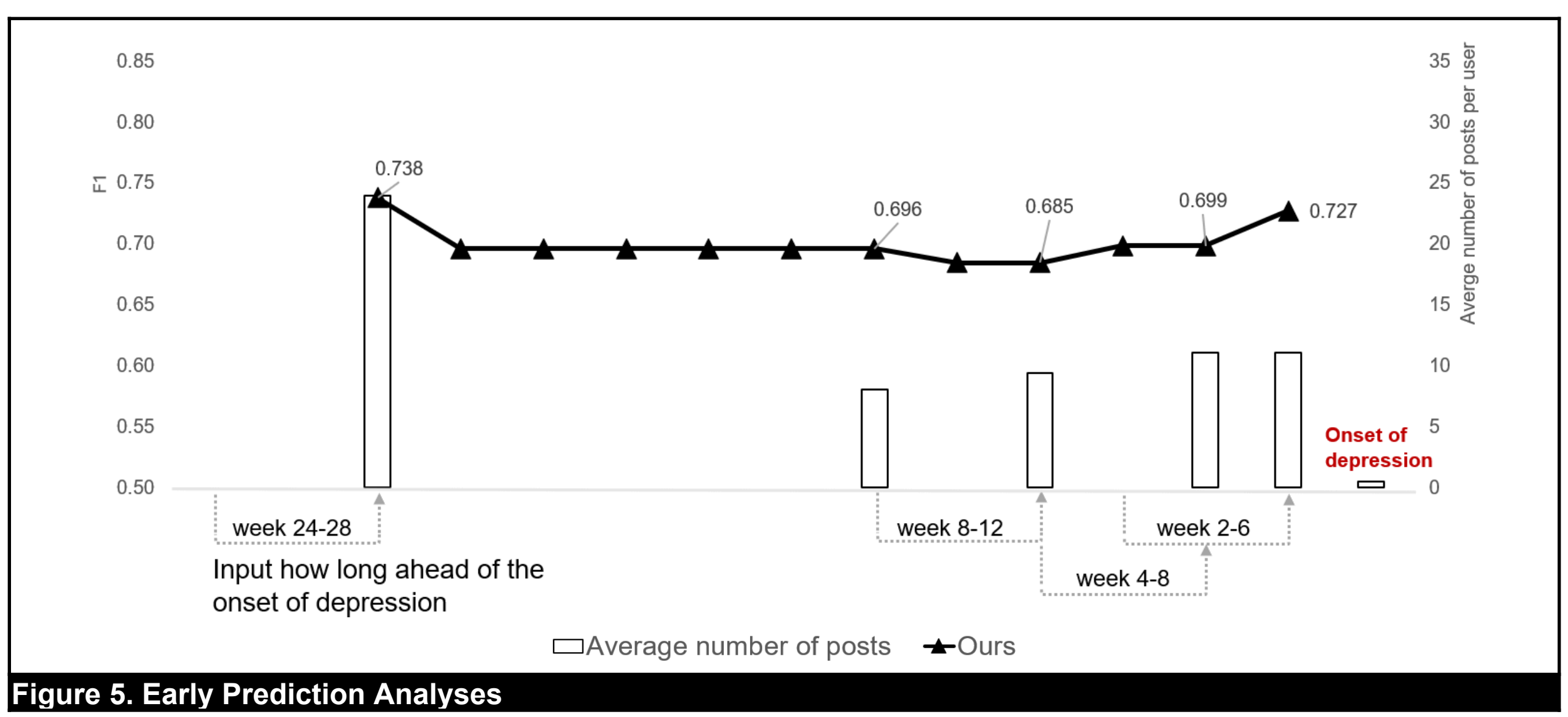

**Figure 5. Early Prediction Analyses**

We also examine how far back in time our framework's input needs to be traced regarding the prediction. For real-world applications, utilizing the user's entire historical data for prediction can result in lengthy sequences and substantial data, posing challenges in data gathering and storage. We utilize data from $x$ weeks before the prediction until the prediction time as input ( $x = \{2, 4, 8, 24\}$). The challenge in this prediction task also lies in that, as the input period shortens, the number of posts decreases, and signals related to depression also weaken. The experimental results (Figure 5) show that the performance of our method is not significantly affected by the shortened time window. It achieves F1 scores over 0.8 using data 12-24 weeks (i.e., 3-6 months) before the prediction (F1 = 0.826) and it shows an F1 score close to 0.8 even when using only two weeks' data (F1 = 0.793). These results indicate that our method can yield a satisfactory prediction performance when there is data available from a specific period before the prediction rather than the entire historical data, enhancing the applicability of our method.

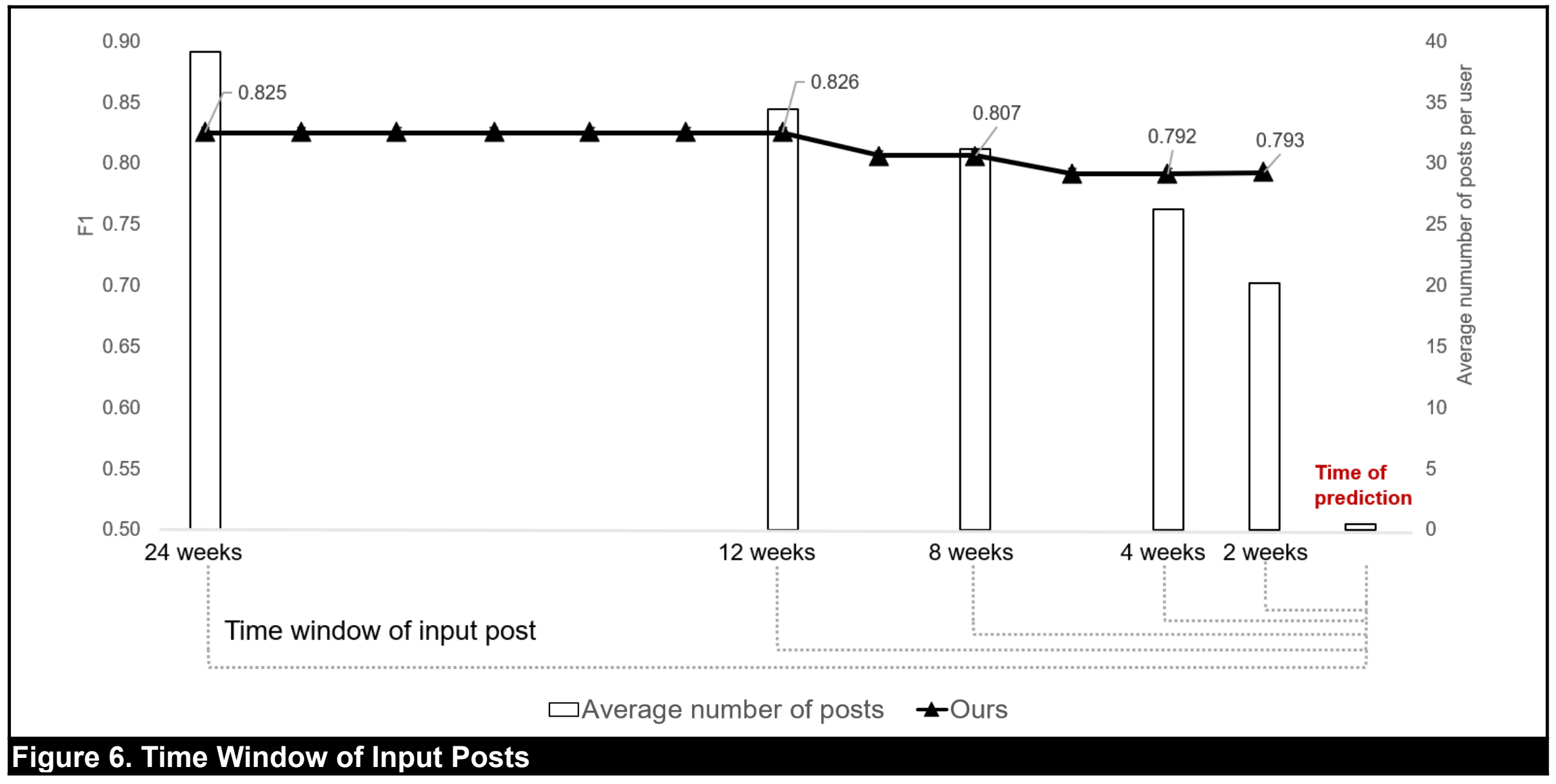

**Figure 6. Time Window of Input Posts**

### 4.4.3. Generalizability to Rare Disease

We further demonstrate how our method, leveraging its few-shot learning capability, can be generalized to emerging or rare diseases using the *self-harm* dataset from the eRisk database, where labeled positive data records are scarce (i.e., 41 subjects).

**Table 8. Rare Disease Dataset Summary**

| Dataset | | # of subjects | # of posts | # of words | Avg # of posts per subject | Avg # of days from first to last post |
|---|---|---|---|---|---|---|
| Self-harm | *P* | **41** | 6,927 | 171,789 | 169 | 495 |
| | *N* | 299 | 163,506 | 3,073,912 | 546 | 500 |

*Note:* P = positive examples; N = negative examples.

We report the evaluation results in comparison to other studies on mental disorder detection (Table 9). Our method consistently outperforms others in terms of F1 score (ours: 0.917 vs. the best traditional machine learning with feature engineering: 0.755 (Reece et al., 2017), and the best deep learning with representation learning: 0.715 (Malviya et al., 2021).

We also conducted ablation studies to validate the contribution of each component in our ensemble prompt method to the prediction results. Removing either prefix tuning or the rule-based prompt significantly degrades performance in terms of F1 score (0.917 → 0.849 and 0.917 → 0.886, respectively). Removing both components, relying solely on the LLM (Flan-T5), leads to an even greater performance drop (0.917 → 0.789). These ablation studies demonstrate that our ensemble prompt method effectively leverages prefix tuning to capture personalized context, even when the number of labeled samples is extremely limited. Additionally, rule-based tuning ensures the stable incorporation of existing medical knowledge, thereby maximizing the LLM's capabilities and enhancing its ability to detect mental disorders using user-generated content.

**Table 9. Comparison with Other Methods on Rare Disease Dataset**

| | | Model | AUC | F1 | Precision | Recall |
|---|---|---|---|---|---|---|
| Other Mental Disorder Detection Studies | Traditional Machine Learning with Feature Engineering | Choudhury et al. (2013) | 0.739 ± 0.003 | 0.710 ± 0.002 | 0.690 ± 0.005 | 0.739 ± 0.003 |
| | | Coppersmith et al. (2014) | 0.711 ± 0.005 | 0.727 ± 0.001 | 0.748 ± 0.002 | 0.711 ± 0.001 |
| | | Preoţiuc-Pietro et al. (2015) | 0.789 ± 0.007 | 0.671 ± 0.002 | 0.647 ± 0.001 | 0.789 ± 0.003 |
| | | Benton et al. (2017) | 0.639 ± 0.024 | 0.662 ± 0.018 | 0.725 ± 0.003 | 0.639 ± 0.024 |
| | | Reece et al. (2017) | 0.783 ± 0.071 | 0.755 ± 0.027 | 0.746 ± 0.002 | 0.783 ± 0.071 |
| | | Chau et al. (2020) | 0.847 ± 0.001 | 0.748 ± 0.001 | 0.708 ± 0.008 | 0.847 ± 0.001 |
| | Deep Learning with Representation Learning | CNN-based (Lin et al. 2020) | 0.699 ± 0.043 | 0.653 ± 0.042 | 0.778 ± 0.075 | 0.563 ± 0.023 |
| | | LSTM-based (Khan et al. 2021) | 0.534 ± 0.012 | 0.582 ± 0.054 | 0.529 ± 0.006 | 0.668 ± 0.146 |
| | | Transformer-based (Malviya et al. 2021) | 0.761 ± 0.032 | 0.715 ± 0.041 | 0.883 ± 0.035 | 0.601 ± 0.042 |
| Ablation Studies | LLM with Prompt Engineering | Flan-T5 | 0.806 ± 0.008 | 0.789 ± 0.003 | 0.723 ± 0.005 | 0.868 ± 0.002 |
| | | Flan-T5 + Prefix Tuning | 0.892 ± 0.000 | 0.886 ± 0.002 | 0.840 ± 0.009 | 0.936 ± 0.007 |
| | | Flan-T5 + Rule-based prompt | 0.833 ± 0.003 | 0.849 ± 0.003 | 0.939 ± 0.004 | 0.775 ± 0.000 |
| | | Flan-T5 + Prefix Tuning + Rule-based prompt **(Ours)** | **0.924 ± 0.000** | **0.917 ± 0.008** | 0.848 ± 0.001 | 1.000 ± 0.000 |

## 4.5. Explaining Prediction Results

To explain our prediction results and demonstrate how our prompt engineering method enhances model performance, we conduct two post hoc analyses.

In the first analysis, we randomly select the predicted probabilities for certain segments of user-generated content (i.e., denoted as $\bar{x}$ in our method). These probabilities represent our method's assessment of the relevance of $\bar{x}$ to a particular mental disorder. We present the content of $\bar{x}$ and the corresponding probabilities as bar charts in Figure 7. For example, in the case of depression, the model assigns a relevance score of 0.9916 to the statement "*I don't have many friends.*" Our model not only can determines the relevance between $\bar{x}$ and a mental disorder but also identifies whether $\bar{x}$ represents symptoms, life events that may exacerbate the disorder, or whether a social media user is self-reporting ongoing treatment. To validate whether our method's judgment of these statements aligns with medical professionals' evaluations, we invited two psychiatrists (both having an MD and more than 4 years of clinical experience) to review the randomly selected set of $\bar{x}$. Each psychiatrist independently rated these $\bar{x}$ based on their medical expertise. The ratings were on a scale from 1 to 5: a score of 1 indicated that the content was unlikely to have been made by a person with a particular mental disorder, while a score of 5 indicated that the content was highly likely to have been made by a person with a particular mental disorder. We used the Pearson correlation coefficient to measure the inter-rater reliability between the two psychiatrists: their assessments showed a high level of agreement (p = 0.9636) (Table 10). Next, we evaluated the alignment between the ratings of two psychiatrists and the predictions generated by our model. The Pearson correlation coefficients indicate that the correlation between our method's results and Expert 1's ratings is 0.8467, while the correlation with Expert 2's ratings is 0.8688. Additionally, the correlation between our method's results and the average psychiatrists' ratings is 0.8656. These findings demonstrate that our method closely aligns with expert assessments in detecting the relevance of user-generated content to a particular mental disorder, thereby confirming the reliability of our approach.

In practical applications, the identified set of $\bar{x}$ with high probabilities can be provided to stakeholders (e.g., medical professionals or platforms), along with an explanation of why a social media user may be at risk for a particular mental disorder, thereby enabling appropriate interventions. Over a longer time period and at the population level, a collection of $\bar{x}$ with high probabilities can be aggregated, enabling researchers to examine mental disorder status across different platforms or regions with varying user profiles. This analysis can reveal the most common mental health issues (e.g., symptoms) and track changes in the prevalence of mental disorders within a population group (e.g., changes in symptoms before and after major events such as the COVID-19 pandemic, presidential elections, or wars). Such capabilities cannot be achieved with simple discrete prompts using LLMs, as these models cannot provide probabilities or explain why they identify a user as having a mental disorder.

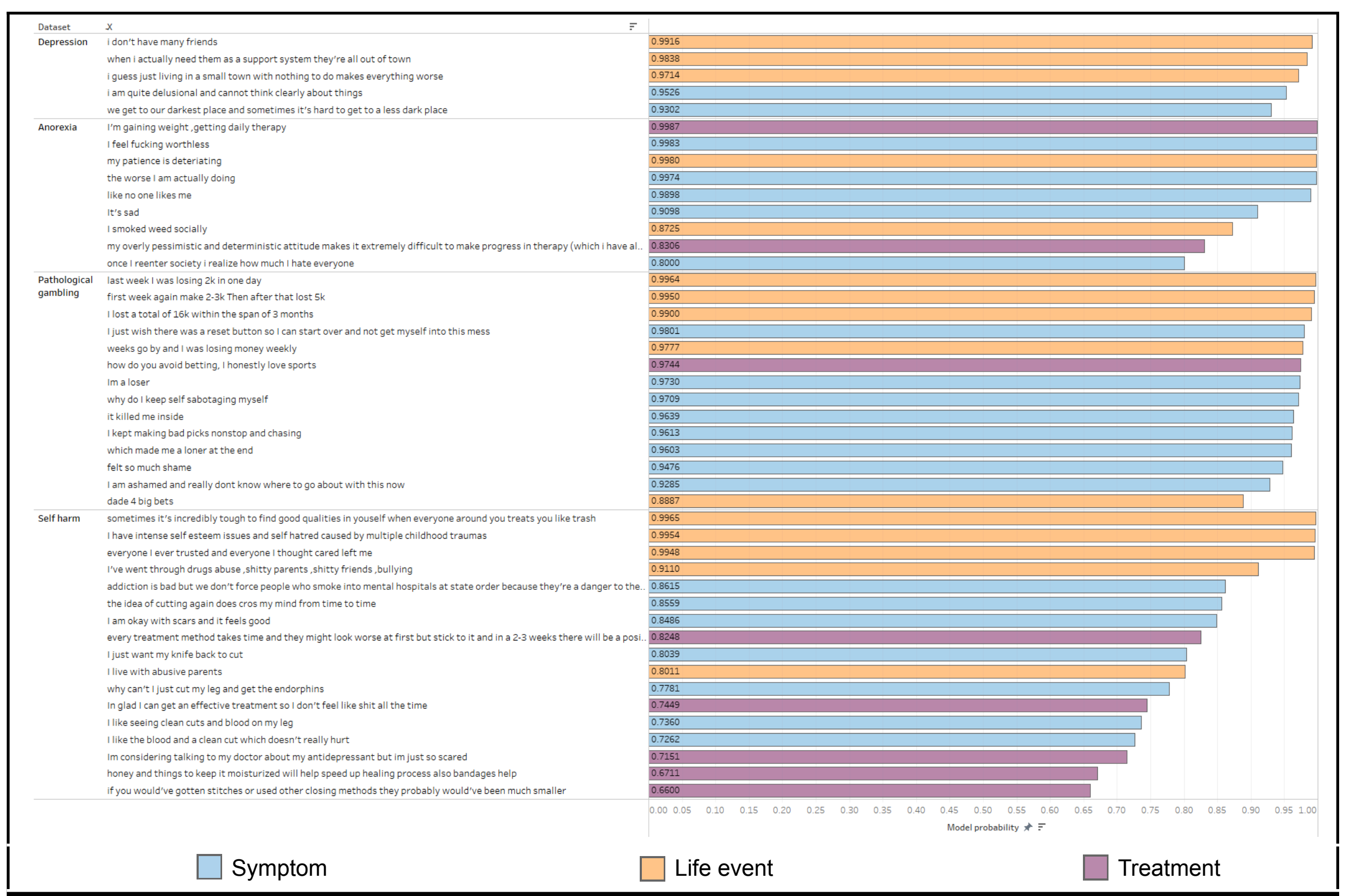

**Figure 7. Model's Predicted Probability of a Post Being Related to a Mental Disorder**

**Table 10. Alignment Between Model Prediction and Expert Ratings**

| | **Pearson correlation coefficient** |
|---|---|
| **Inter-rater reliability[a]** | 0.9636 |

| | **Pearson correlation coefficient** | | |
|---|---|---|---|
| | *Prediction VS Experts average* | *Prediction VS Expert 1[c]* | *Prediction VS Expert 2[d]* |
| **Alignment of prediction and expert ratings[b]** | 0.8656 | 0.8467 | 0.8688 |

*Note:* [a] Pearson’s correlation coefficient is used to measure the pairwise correlation between the two raters. The rating scale ranges from 1 to 5, with 5 indicating the highest relevance to a disorder and 1 indicating the lowest. The rating scale is ordinal and assumed to be continuous.
[b] Pearson’s correlation coefficient is used to measure the alignment between the model's predicted probability of a post being related to the disorder and the expert ratings.
[c] The rater is a psychiatrist with an MD and four years of clinical experience.
[d] The rater is a psychiatrist with an MD, a PhD, and eight years of clinical experience.

In the second analysis, we visualize and compare the vector representations of both the original user-generated content and the content transformed by our prompt method. The visualization results are presented in Figure 8. Through visualization, we can easily observe that the vector representations of user-generated content for both positive and negative examples (with or without a mental disorder) are intermingled. This aligns with our intuition, as user-generated content online typically covers a wide range of topics, with only a small portion related to mental disorders. Consequently, mental disorder detection faces a high level of noise. By contrast, when we visualize the vector representations of user-generated content processed through our prompt method, we observe a significantly increased separation between positive and negative examples. For the task of mental disorder detection, our prompt engineering method effectively reduces the complexity of classification/detection. This analysis explains why our prompt engineering method optimizes LLM performance and improves prediction accuracy.

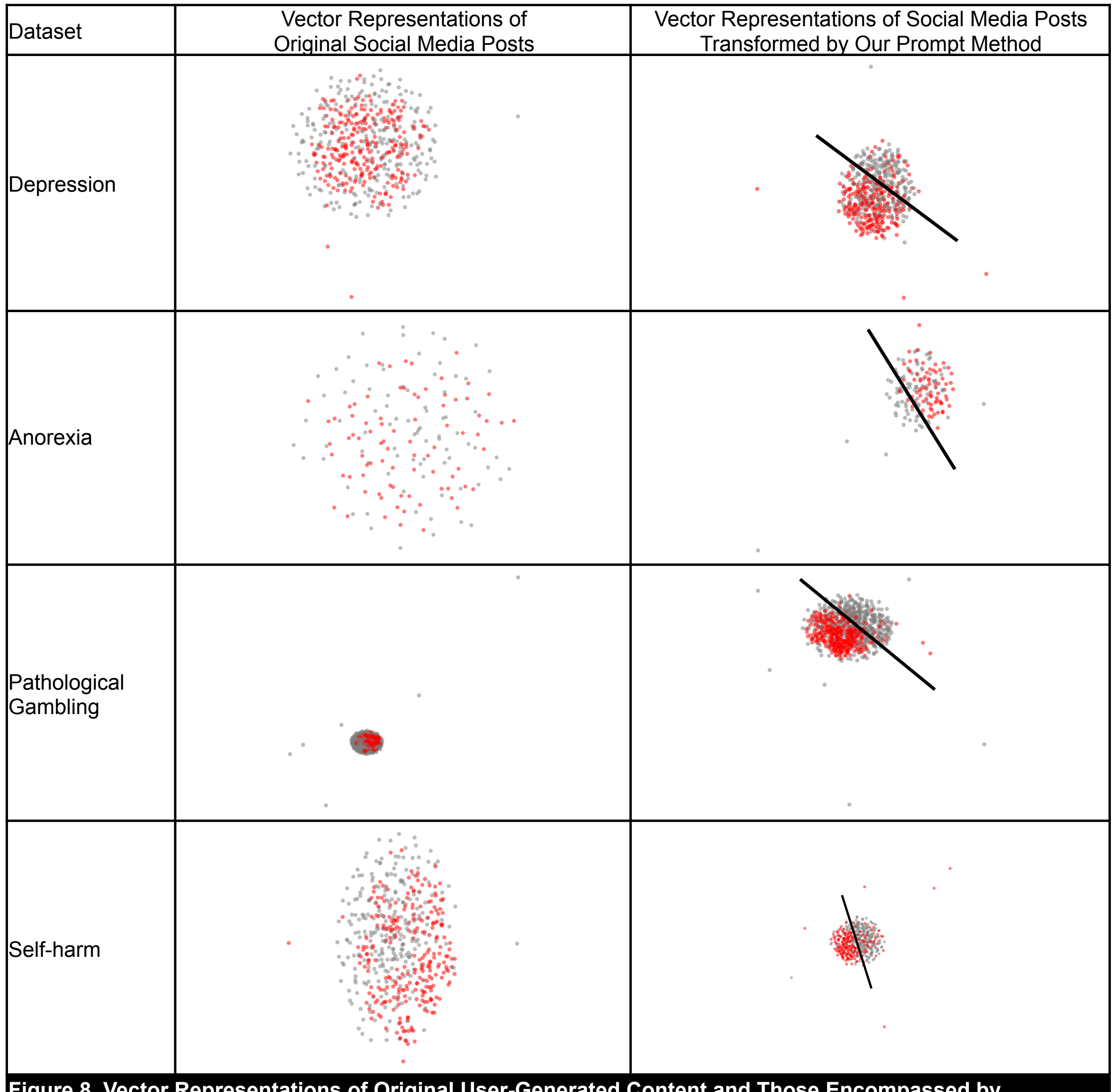

**Figure 8. Vector Representations of Original User-Generated Content and Those Encompassed by Continuous Prompting**

*Note:* • P = positive examples; • N = negative examples.
The vector representations are visualized using t-SNE (Maaten and Hinton 2008).

# 5. DISCUSSION

In this work, we adapt advanced AI techniques, including large language models and prompt engineering, to explore how AI can contribute to mental disorder detection on user-generated textual content. The significant advantages of our research lie in its pioneering approach to overcoming long-standing barriers in this field: eliminating the need to collect large amounts of labeled training data for each specific disease or to design specialized supervised learning model architectures for each research problem. At the same time, we place a strong emphasis on addressing the technical challenges using user-generated text data in the healthcare domain. This includes developing personalized methods to represent each user's uniqueness, as well as seamlessly integrating and leveraging disease-related medical knowledge to provide context for the task, instruct learning objectives, and operationalize prediction goals for various types of mental disorders.

We evaluate the effectiveness of our research design by employing multiple mental disorders as research cases. The experimental results reveal several key findings. First, the performance of our approach, which combines prompt engineering with LLMs, significantly outperforms other supervised learning paradigms, including feature engineering and architecture engineering. While LLMs inherently are more powerful than many supervised learning models, even using the same LLM as the base model, our designed prompt engineering approach still outperforms alternative prompting strategies, which is the main focus of our work. Second, our method successfully accomplishes few-shot learning for various mental disorders. In other words, we only need to provide the AI model with a small number of examples, and it can effectively identify potential users at risk of mental disorders within user-generated textual content. Third, by breaking down the original complex prediction task into subtasks and intentionally incorporating medical knowledge into prompt engineering, our approach not only significantly reduces the prediction difficulty and improves the performance of LLMs but also enhances the interpretability of our prediction results, making it easier for stakeholders to understand and utilize the prediction results.

From a design science perspective, our contributions are threefold. First, we introduce a novel framework rooted in LLMs and prompt engineering, which enables the few-shot detection of multiple mental disorders through user-generated text content. Notably, it bestows a significant advantage by eliminating the need for an extensive volume of labeled training data or the intricate engineering of customized architectures for each distinct disease or research problem. Second, within our framework, we employ a multi-prompt engineering approach, effectively synergizing various prompt engineering techniques—specifically, prefix tuning and rule-based prompt engineering. This strategy is tailored to tackle the distinctive technical challenges within the healthcare domain, entailing the use of personalized prompts and the integration of existing medical domain knowledge, which significantly elevates the accuracy and efficacy of our methods. Third, as part of our framework, we propose a new rule-based prompt engineering method, which efficiently breaks down complex textual content-based detection problems and seamlessly integrates domain knowledge existing in ontology format (one of the widely adopted formats for domain knowledge). The versatility of this new rule-based prompt engineering method extends to other research problems that require the decomposition of challenging tasks and can maximize the utilization of LLM's potential to address real-world issues.

In addition to technical contributions, our study has implications for IS such as computational design science, healthcare IS, and business analytics and intelligence. Our research also has significant practical implications. We have demonstrated the application of prompt engineering and the utilization of LLMs in the realm of detecting mental disorders through user-generated text content. User-generated content platforms and medical professionals can utilize similar methods to incorporate medical knowledge, enabling the large-scale discovery and monitoring of other chronic diseases in a low-cost way, which can facilitate early identification and intervention of chronic diseases. With minor adjustments to the applied medical knowledge, our approach can also be expanded to the management of more chronic diseases. Taking diabetes as an example, we can adapt the medical domain knowledge to observe, track, and intervene in users' lifestyles, diets, physical activity, and sleep patterns; we can not only identify chronic diseases but also improve their management and potentially prevent them. Ultimately, this can enhance the well-being of individuals with chronic conditions, control healthcare costs, and ensure a more efficient and accessible healthcare system.

Our work comes with limitations. First, the ethical and privacy considerations. The implementation of AI-based technology in healthcare necessitates a careful consideration of ethical and privacy concerns. It is imperative to ensure the responsible and ethical use of AI in the healthcare domain, safeguarding the privacy and well-being of patients and individuals. Our approach takes into account personalized prompts, which can to some extent address the issue of user privacy. This direction is still worthy of further research. Second, hallucination issues with LLMs can arise (e.g., fake, random, or irresponsible responses from LLMs). In healthcare applications, where accuracy in detection is of utmost importance, the need for careful scrutiny of detection results cannot be overstated. By integrating medical knowledge into our work, we have substantially reduced the occurrence of such issues. Exploring how the introduction of domain-specific knowledge can effectively curb irresponsible responses from LLMs is a promising avenue for future research. Another future research direction involves the exploration of multimodal large pre-trained models capable of processing text, images, or videos. Our framework can be extended to user-generated multimodal data, encompassing images and videos that capture users' body shapes, food consumption, exercises, living environments, and more. Consequently, this extension enables the identification of chronic diseases associated with more lifestyle factors, which may be implicitly manifested in images and videos. Such insights can be harnessed for the detection and management of chronic diseases.

## 6. CONCLUSION

We explore whether state-of-the-art AI technologies can make a difference in chronic disease management within the context of user-generated content. Our study demonstrates the immense potential and promise of AI in the healthcare domain: it not only enhances chronic disease detection accuracy but also reduces the need for a large number of labels in machine learning and the necessity for architecture engineering for each specific research question.

AI's capacity to perform tasks effectively holds great potential for enhancing healthcare management. AI offers the advantages of (1) tailoring solutions to specific healthcare needs, enabling customization, and (2) automating a multitude of tasks, alleviating the workload on healthcare professionals and enhancing overall efficiency. The integration of AI into healthcare management has the potential to significantly enhance the precision and efficiency of administrative processes. By optimizing chronic disease management and streamlining healthcare procedures, AI can facilitate interventions and more effective treatments for patients. Through increased efficiency and early disease recognition, AI can aid healthcare institutions in cost savings—an especially noteworthy benefit, given the resource-intensive nature of the healthcare sector, with any funds saved being available for reinvestment in the enhancement of patient care.

## APPENDIX 1

**Depression Ontology Evaluation.** In this work, we construct mental disorder ontologies as the medical knowledge base of our proposed method. To evaluate the quality of the ontologies, we leverage a medical expert panel and adapt previous work in evaluations of ontologies (Ouyang et al. 2011) and measure the coverage of the ontologies by comparing the number of concepts in a ontology with regard to widely used mental disorder diagnosis scales, including DSM-5-TR Self-Rated Level 1 Cross-Cutting Symptom Measure - Adult (DSM-5-TR) (APA 2022), Patient Health Questionnaire (PHQ-9) (Martin et al. 2006), and Quick Inventory of Depressive Symptomatology-Self-Report (QIDS-SR) (Rush et al. 2003). For the medical expert panel, we invite two psychiatrists from a nationally recognized hospital to review our mental disorder ontologies. After the initial independent review, the psychiatrists met, added entities that are missing from the initial version, and removed entities that are redundant and not clinically relevant. We use the final ontologies verified by the medical experts in our study. Specifically, we define $C_j = \{c_1, c_2, \ldots c_i, \ldots c_n\}$ as the set of $n$ concepts in the disease $d_j$'s ontology $O_j$, which includes $d_j$'s symptoms, major life events, and treatments. Let $T = \{t_1, t_2, \ldots t_j, \ldots, t_m\}$ be the set of $m$ medical terminology in the diagnosis criteria $D$. The coverage of $O_j$ to $D$ is calculated as

$\frac{\sum_{c \in c(O)} \sum_{t \in T} I(c,t)}{m}$; if there is a $t_k$ or $t_k'$ synonyms in $\{c_1, c_2, \ldots, c_n\}$, set $I(c, t) = 1$, otherwise, set $I(c, t) = 0$. The resulting coverages of the ontologies to DSM-5-TR, PHQ-9, and QIDS-SR are presented in Table A1.

**Table A1. The Coverage Rate of Ontologies to Mental Disorder Diagnosis Scales**

| Mental disorder diagnosis scales | | DSM-5-TR | PHQ-9 | QIDS-SR |
|---|---|---|---|---|
| Coverage rate | Depression | 85.6% | 95.2% | 93.8% |
| | Anorexia | 84.8% | – | – |
| | Pathological gambling | 96.3% | – | – |
| | Self-harm | 93.5% | – | – |

The medical terminologies such as “angry” and “little energy” are not included in the ontologies. These terminologies are commonly used in populations other than mental disorders, and therefore, they are not suitable for determining whether a person has a mental disorder outside the context of using the mental disorder diagnosis scales. Overall, the coverage calculation results demonstrate that our ontology can comprehensively cover the widely used depression scales, indicating that it is suitable as the knowledge base for the proposed method.